\documentclass[
aps,
physrev,
superscriptaddress,
]{revtex4-2}

\usepackage{graphicx}
\usepackage{dcolumn}
\usepackage{bm}
\usepackage{hyperref}
\usepackage{url}

\usepackage{booktabs}       
\usepackage{amsfonts}       
\usepackage{nicefrac}       
\usepackage{microtype}      
\usepackage{xcolor}         

\usepackage{amsmath}
\usepackage{amssymb}
\usepackage{mathtools}
\usepackage{amsthm}
\usepackage{tabularx}

\usepackage[utf8]{inputenc}

\usepackage[capitalize,noabbrev]{cleveref}

\newcommand{\lrpar}[1]{\left(#1\right)}
\newcommand{\lrbra}[1]{\left[#1\right]}

\begin{document}
\preprint{APS/123-QED}

\title{Characterizing Nonlinear Dynamics via Smooth Prototype Equivalences}

\author{Roy Friedman}
\email{Contact author: roy.friedman@mail.huji.ac.il}
\affiliation{School of Computer Science and Engineering, Hebrew University of Jerusalem, Israel;}
\author{Noa Moriel}
\affiliation{School of Computer Science and Engineering, Hebrew University of Jerusalem, Israel;}
\author{Matthew Ricci}
\affiliation{Université Paris Cité, Paris, France;}
\affiliation{Institut Imagine, INSERM UMR 1163, Paris, France;}
\author{Guy Pelc}
\affiliation{School of Computer Science and Engineering, Hebrew University of Jerusalem, Israel;}
\author{Yair Weiss}
\affiliation{School of Computer Science and Engineering, Hebrew University of Jerusalem, Israel;}
\author{Mor Nitzan}
\email{Correspondence: mor.nitzan@mail.huji.ac.il}
\affiliation{School of Computer Science and Engineering, Hebrew University of Jerusalem, Israel;}
\affiliation{Racah Institute of Physics, Hebrew University of Jerusalem, Israel;}
\affiliation{Faculty of Medicine, Hebrew University of Jerusalem, Israel}

%


\begin{abstract}
Characterizing the long term behavior of dynamical systems given limited measurements is a common challenge throughout the physical and biological sciences. This is a challenging task due to the sparsity and noise inherent to empirical observations, as well as the variability of possible long-term dynamics. We address this by introducing smooth prototype equivalences (SPE), a framework for matching sparse observations to prototypical behaviors using invertible neural networks which model smooth phase space deformations.
SPE can localize the invariant sets describing long-term behavior of the observed dynamics through the learned mapping from prototype space to data space.
Furthermore, SPE can classify dynamical regimes by comparing the data residual of the deformed measurements to prototype dynamics. 
Our method outperforms existing techniques in the classification of oscillatory systems and can efficiently identify invariant structures like limit cycles and fixed points in an equation-free manner, even when only a small, noisy subset of the phase space is observed. 
SPE further reveals driving genes in synthetic oscillators such as the repressilator regulatory circuit, and traces cyclic biological processes like the cell cycle trajectory directly from experimental high-dimensional single-cell gene expression data.
\end{abstract}

\maketitle

\section{Introduction}

Single-cell RNA-sequencing provide a static population snapshot of high-dimensional gene expression measurements for thousands of cells simultaneously \citep{lahnemann2020eleven,kolodziejczyk2015technology}. Recent advances enable the estimation of instantaneous changes of gene expression through RNA velocity \cite{la2018rna,bergen2020scvelo}, directly from such static gene expression data. Pairing velocity estimates with scRNA-seq exposes information regarding local velocity in gene expression space of each of the measured cells, yet predicting long-term cellular behavior from these observations remains a challenging problem. 

In stable nonlinear systems, such as those exhibited in some contexts of cellular biology, orbits converge to sets of the phase space which are invariant to the dynamics. The characterization of these invariant sets is of high interest in the context of single-cell biology, where invariant sets in gene expression space correspond to reproducible and stable biological processes, or programs carried out by individual cells. For instance, multiple cellular processes approximately correspond to limit cycles in gene expression space, including the cell cycle, 
where proliferating cells traverse different phases of this cycle at recurring intervals \citep{Schwabe2020-xh,riba2022cycle,karin2023scprisma}, and
the circadian rhythm, where the expression levels of a set of genes oscillate with a diurnal period \citep{dibner2010mammalian,karin2023scprisma}. Conversely, the transition of cells from progenitors to terminally differentiated cells of distinct fates can be modeled as the convergence of cells to multiple node attractors in gene expression space \citep{saez2022cellfate,lange2022cellrank,farrell2023inferring,schiebinger2019optimal}. In this way, the exact form of invariant sets encoded by cells in gene expression space can reveal the set of dynamical processes cells go through in different biological contexts including homeostasis and development, and uncover the genetic programs and interactions that drive these dynamics. Moreover, the forms of deviations from these stable patterns can potentially inform mechanistic interpretations in the context of aging or disease progression. Thus, distinguishing between different possible invariant sets, pinpointing the underlying driving genes and determining the timings of these processes is the subject of intense work in computational biology (e.g. \cite{mahdessian2021fuccicycle,Zheng2022-iw,Kedziora2024-ov,Zinovyev2021-qc,Stallaert2021-xw,qu2025gene,maehara2025geometry}).

In the broader context of data-driven analysis of nonlinear dynamics, substantial work has been devoted to the estimation of vector fields underlying observed time series data in the form of ordinary differential equations (ODEs) \cite{chen2018neural,brunton2016sindy,farrell2023inferring,kochkov2021machine,brunton2017chaos}. However, even when the governing equations are known and the dynamics are stable, the exact structure of long-term equilibrium or oscillatory behaviors is challenging to quantify \citep{hilbert1900, smale2000hilbert}, and this challenge is exacerbated in realistic scenarios given sparse and noisy data, when the governing equations of the dynamics are often unknown apriori. Moreover, invariant sets can exhibit complex geometry in phase space and can change drastically when introduced to small perturbations \citep{guckenheimer2013nonlinear}. These attributes, particularly in high dimensional systems, frequently hinder the detection and localization of such invariant sets \citep{moriel2023let}. 
A different approach, taken by a recent line of work, has sought to determine whether two systems are governed by qualitatively similar dynamics \citep{chen2024dform,bramburger2021deep,redman2022koopman,glaz2024efficient,moriel2023let,ostrow2023beyond} through the use of relaxed notions of dynamical equivalences \citep{Bollt2007relaxing,skufca2008conjugate}. These characterizations are attractive as they focus on the qualitative behaviors of dynamical systems, such as the stability and type of invariant sets. To do so, these approaches compare \emph{two systems}, using observations from long time series in order to form a delay embedding where Takens’ embedding theorem can be employed \citep{takens1981detecting} or to compare spectra of linear approximations to the dynamics in the context of Koopman theory \citep{brunton2021koopman}. It is therefore difficult to apply the above methodologies in practice to many types of experimental data, such as single-cell gene expression data, where cells are destroyed as part of the measurement, and cannot be followed in time. In this work, we offer an alternative view, wherein noisy empirical data is matched to an analytical model guided by domain expertise, regardless of the sparsity, temporal resolution or dimensionality of the observations.

Our work focuses on the identification and tracing of invariant sets in sparse, high dimensional observations measured when the underlying dynamical system is unknown. To extract information about the long-term behavior of the underlying dynamics, we rely on the fact that the topology and stability of the invariant sets of a dynamical system are not affected by a smooth, invertible change of coordinates. Two systems related by such a smooth invertible mapping are called \emph{smoothly equivalent}, and their invariant sets share the same stability and topology \citep{chen2024dform}.
Our approach, called \emph{smooth prototype equivalences} (SPE), learns an invertible mapping, parameterized by an invertible neural network, between the data and a simple, \emph{prototypical} dynamical system. Such a prototypical system could be, for example, a set of governing equations that exhibit a bifurcation between a limit cycle and a node attractor. 


We first demonstrate how SPE can classify and localize node and limit cycle attractors in a synthetic dataset of 2D dynamical systems drawn from models across the physical and life sciences. Furthermore, we show that SPE, unlike previous methods, is robust to noisy and low-sample conditions, illustrating the compatibility of SPE to real data-driven scenarios. We then demonstrate that SPE can naturally extend to higher dimensions by applying it to the synthetic six-dimensional repressilator gene regulatory network \citep{elowitz2000repressilator}.Finally, we illustrate how SPE can be used in a real scientific scenario, by deploying our method to localize  cell-cycle dynamics from noisy, high-dimensional single-cell gene expression measurements, without relying on a gene expression reference atlas.

\section{Method}

\subsection{Smooth prototype equivalences}\label{sec:SPE}

In single-cell RNA sequencing (scRNA-seq) data, individual cells are sequenced and represented by sparse high-dimensional gene expression data. Recent advances in computational biology enable the estimation of local \textit{RNA velocity} directly from scRNA-seq data \cite{la2018rna,bergen2020scvelo}.
Such temporal information can be modeled as arising from a hidden set of ordinary differential equations (ODEs), governing the dynamics of the whole biological system.
In this work, we assume access to such data, corresponding to sparse observations from a subset of the phase space of an ODE.
Mathematically, let $\mathcal{D}=\left\{\lrpar{x_i,\dot{x}_i}\right\}_{i=1}^N$ be a dataset of $N$ pairs of positions $x_i\in\mathbb{R}^d$ and their respective velocities $\dot{x}_i\in\mathbb{R}^d$, which were generated by a \emph{hidden} ground-truth structurally stable ODE, $\dot{x}=f(x)$. 
Realistically, observed data represents noisy estimates of the velocities, $\dot{x}_i=f(x_i)+\epsilon_i$, where $\epsilon_i$ is a per-sample noise term.

\begin{figure*}[!tbh]
\begin{center}
    \includegraphics[width=\linewidth]{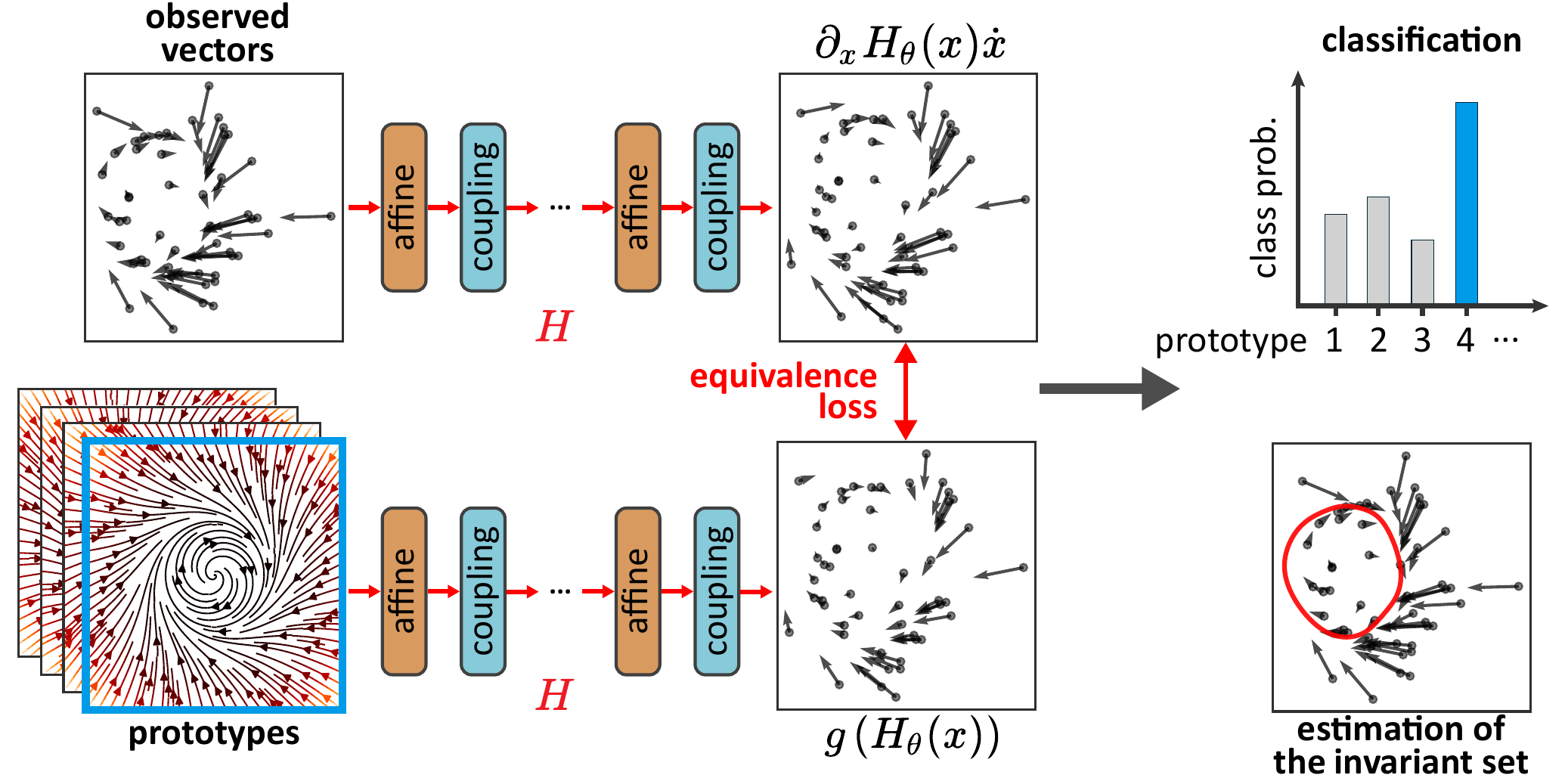}
\end{center}
  \caption{\textbf{Fitting dynamical prototypes using \emph{smooth prototype equivalences} (SPE).} The observed dynamics (positions $x_i$ and velocities $\dot{x}_i$) are compared to each of the chosen prototypes, $g(y)$, which involves training an invertible neural network (INN) $H_\theta(x)$ for each prototype $g(y)$. After training, the observed dynamics are classified as the prototype that achieves the smallest equivalence loss (\Cref{eq:diffeo-loss}). A trained INN $H_\theta(x)$ provides a mapping between data space and prototype space, which in turn can be used to estimate the long-term behavior, i.e. invariant set, of the dynamics underlying the observed data.}
  \label{fig:schematic}

\end{figure*}

To effectively restrict the search space when modeling the dynamics of the observed data, we use strong prior information for the possible dynamical patterns within  the data.
In other words, we can constrain our model to only match dynamical systems similar to a candidate set of governing equations, $\dot{y}=g(y)$. In this scenario, the dynamics of $g(y)$ form a \emph{prototype} for the behavior expected to exist in the observed data.

To achieve this, we utilize the notion of \emph{smooth equivalences} between dynamical systems \citep{chen2024dform}. Specifically, two systems $\dot{x}=f(x)$ and $\dot{y}=g(y)$ are said to be smoothly equivalent if:
\begin{equation}\label{eq:smooth-equivalence}
    \exists H\ \ \text{s.t.}\ \ \partial_x H(x)\ \dot{x}=g\lrpar{H(x)}
\end{equation}
where $H:\mathcal{X}\rightarrow \mathcal{Y}$ is a diffeomorphism and $\partial_x H(x)$ is the abbreviated notation for the Jacobian of $H$ with respect to $x$. If two systems obey this equivalence relation, each orbit of the system $f(x)$ has a smooth one-to-one mapping to an orbit in $g(y)$, and thus the two systems share the same qualitative behavior, such as the topologies and stabilities of their invariant sets. 
In this work, we aim to learn a diffeomorphism between the data, $\dot{x}=f(x)$, and the prototype, $\dot{y}=g(y)$, and
we parametrize such a diffeomorphism using an invertible neural network (INN), $H_\theta(x)$, which is optimized in order to approximately satisfy \Cref{eq:smooth-equivalence}. This is achieved by minimizing the following \emph{equivalence loss} \citep{chen2024dform}:
\begin{equation}
    L_\text{E}(H_\theta, g) = \frac{1}{N}\sum_{i=1}^N \left\|\frac{\partial_{x_i} H_\theta(x_i)\dot{x}_i}{\|\partial_{x_i} H_\theta(x_i)\dot{x}_i\|}-\frac{g\left(H_\theta(x_i)\right)}{\| g\left(H_\theta(x_i)\right)\|}\right\|^2 \label{eq:diffeo-loss}    
\end{equation}
When the equivalence loss is 0, the dynamics that are generated by $\dot{x}$ and $g(y)$ are smoothly equivalent for any point $x\in\mathcal{D}$. 

In practical scenarios given sparse and noisy observations, the loss in \Cref{eq:diffeo-loss} will not necessarily be equal to 0 even when the two underlying systems are equivalent. Nonetheless, the equivalence loss provides a good indication of similarity between dynamical systems when there is access to a large amount of samples \citep{chen2024dform}. To improve convergence, we optimize the diffeomorphism using a regularized version of \Cref{eq:diffeo-loss}, penalizing transformations with a large determinant and those that push the prototype's invariant set far from the data (see \ref{app:opt-criteria} for full details).

\subsection{Invariant set localization and prototype classification with SPE}

Our explicit construction using a mapping between the prototype and the observed data allows us to effectively \emph{localize} structures in the observed data, which are in general very difficult to find \citep{smale2000hilbert}. For example, suppose that our prototype $g(y)$ has a limit cycle, and we can generate points on this limit cycle $\gamma=\left\{y_1,\cdots,y_M\right\}$. Given the learned explicit mapping between the two spaces $H_\theta$, $H_\theta^{-1}\lrpar{\gamma}=\left\{H_\theta^{-1}(y_1),\cdots,H_\theta^{-1}(y_M)\right\}$ is guaranteed to be a closed loop in data space. Moreover, if the equivalence loss between the observed data and the prototype is low, then $H_\theta^{-1}\lrpar{\gamma}$ will be a set of points close to the limit cycle of the true hidden system, $\dot{x}=f(x)$. 

Moreover, SPE allows for classification between a set of prototypes, as the equivalence loss directly corresponds to a measure of dissimilarity. So, given a dictionary of possible prototypes $\left\{g_k(y)\right\}_{k=1}^K$, we optimize a diffeomorphism $H_k(x)$ for each, using the equivalence loss from \Cref{eq:diffeo-loss}. We then calculate the equivalence loss, $\text{L}_\text{E}(H_k,g_k)$, between the observed data and each of the prototypes. The data is then classified according to the prototype that has the smallest equivalence loss.

\subsection{Modeling diffeomorphisms with invertible neural networks}\label{sec:nf-diffeos} 

The equivalence loss from \Cref{eq:diffeo-loss} acts as a training objective for fitting the diffeomorphism between data and prototype space, $H_\theta$. To properly capture the mapping between data and prototype space, $H_\theta$ has to be expressive enough to capture the correct mapping while also being computationally lightweight. Standard neural networks (NNs) are apt choices as tunable parametric functions for such tasks, but are not guaranteed to be invertible, which is one of our core requirements from $H_\theta$. 
Instead, we opt to use \emph{invertible neural networks} (INNs), a specific type of NNs which are adapted from normalizing flows (NFs, \cite{papamakarios2021normalizing}), to model our parametric diffeomorphisms.
In INNs, each layer is invertible by design, which makes inverting the whole mapping a simple algorithmic procedure. Furthermore, INNs enable efficient and explicit computations of both the log-determinants of the Jacobian as well as the Jacobian-vector products (JVPs), $\partial_x H_\theta(x) \dot{x}$, needed to calculate the equivalence loss. As such, INNs are particularly well suited as the basis of our parametric diffeomorphisms.

Our INNs utilize a small number of blocks, composed of alternating invertible linear transformations and affine coupling layers (\cite{dinh2014nice}, see \Cref{app:diff-details} for more information). Both of these layer types are adapted so that their JVPs can be calculated in closed form. Specifically, instead of using multi-layer perceptrons (MLPs) typically found in affine coupling layers \citep{papamakarios2021normalizing} we use cosine features, which we term \emph{Fourier feature coupling}. These Fourier feature coupling layers are highly expressive even when the network is narrow and shallow, and their JVPs can be calculated efficiently and in closed form. 
For more details regarding network design choices, see \Cref{app:diff-details}.

\section{Results}\label{sec:results}
\subsection{Reconstructing attractors from sparse data}
\label{sec:recon}
\begin{figure*}[!tbh]
\begin{center}
    \includegraphics[width=\linewidth]{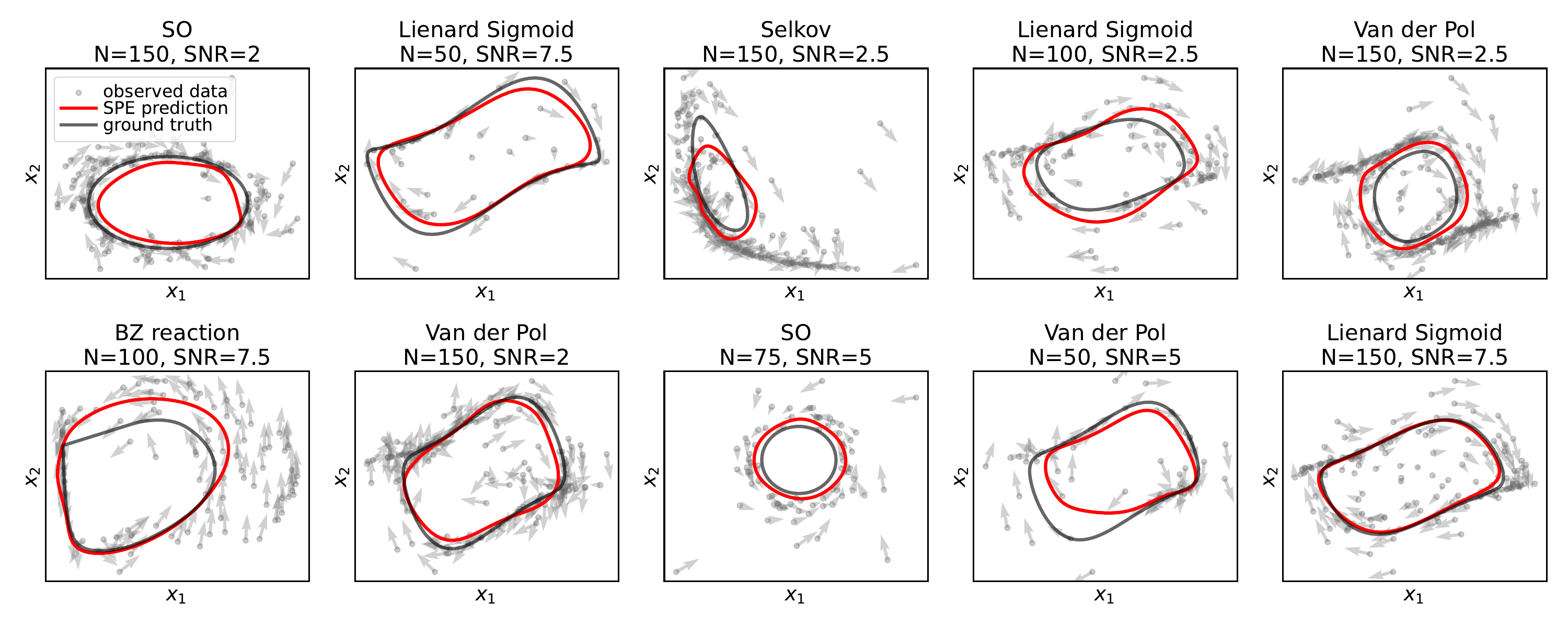}
\end{center}
  \caption{\textbf{Predicting limit cycles for various 2D dynamical systems  using SPE.}
  Examples of invariant sets predicted by SPE (red) from observed vectors (gray), for different dynamical systems which exhibit a limit cycle attractor. Ground-truth trajectories were simulated for each hidden system (black) for visual comparison. 
  Dynamical systems: simple oscillator (SO), BZ reaction, Sel\'kov, Li\'enard sigmoid, Li\'enard polynomial and Van der Pol); corresponding order parameters were uniformly chosen from the set of ranges that exhibit a limit cycle  (\Cref{app:2D-systems}).
  }
  \label{fig:invariants}
\end{figure*}

Locating attractors given realistic observations is generally a challenging problem since sparsity and noise make the underlying, unknown dynamical equations strongly undetermined. We demonstrate how SPE circumvents this difficulty on data simulated from a set of standard families of two-dimensional dynamical systems exhibiting limit cycles. The governing equations for these families of systems can be found in \Cref{app:2D-systems}.

To that end, we optimize a transformation between observed systems and a prototype chosen to flexibly model point and oscillatory dynamics. We use the simple oscillator system, described by a set of a coupled pair of ODEs, defined in polar coordinates by:
\begin{equation}
    \label{eq:prot}
    \dot{r} = r(a - r^2),\qquad
    \dot{\theta} = \omega
\end{equation} 
Here, $a$ is an order parameter that controls whether the system has a node attractor (negative values) or a limit cycle (positive values), and $\omega$ determines the orientation of the cycle (clockwise when negative and counter-clockwise when positive). Data were acquired by sampling initial conditions and points along simulated trajectories (details in \Cref{app:sparse-sim}). 

After fitting an INN $H_\theta(x)$ between the observed data $\mathcal{D}$ and our limit cycle prototype, we estimated the location of the limit cycle in data space by uniformly sampling a sequence of points, $T_\mathcal{Y}= \{y_j\}_{j=1}^M$, which lie along the limit cycle in prototype space, $\mathcal{Y}$. Each point was then mapped back into data space to define a trajectory along the estimated invariant set of the observed dynamics: $\hat{T}_\mathcal{X}=H_\theta^{-1}(T_\mathcal{Y})=\left\{H_\theta^{-1}(y_j)\right\}_{j=1}^M$. 
SPE can accurately trace the structure of limit cycles across diverse dynamical systems given sparse and noisy observations (\Cref{fig:invariants} and Appendix \Cref{fig:more-examples}).

\begin{figure}[!tbh]
\begin{center}
    \includegraphics[width=\linewidth]{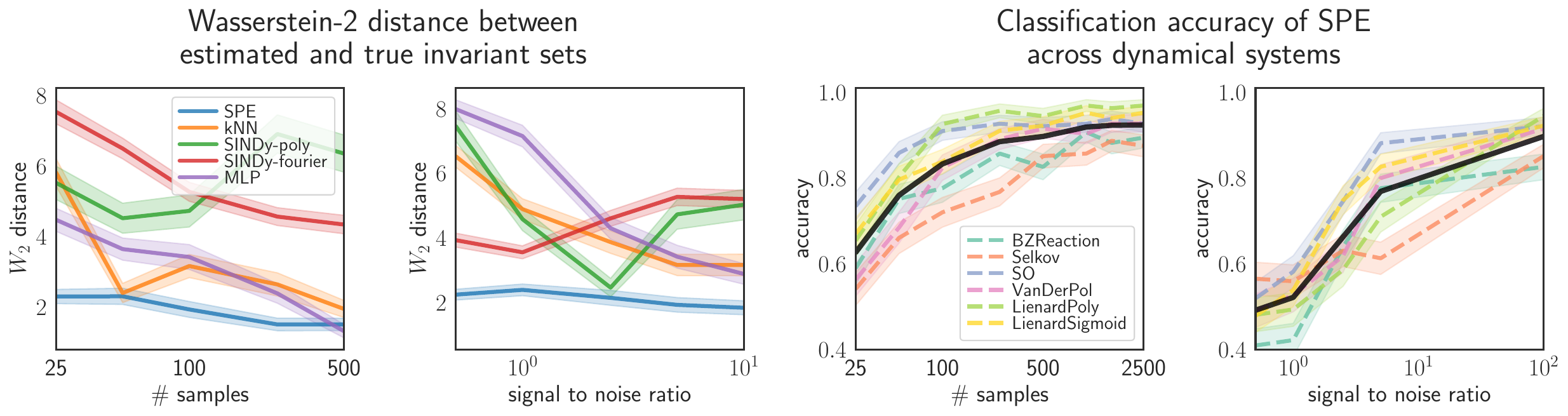}
\end{center}
  \caption{\textbf{
  Evaluation of invariant sets prediction across synthetic dynamical systems.
  } 
  \textbf{Left:} Wasserstein (W2) distance between the estimated and true invariant sets, averaged over 1000 randomly sampled datasets (see \Cref{app:invariant-eval}), produced by SPE along with additional baselines ($k$-nearest neighbor interpolation (kNN), SINDy \cite{brunton2016sindy} using polynomial basis functions (SINDy-poly), SINDy using Fourier features (SINDy-fourier), and a multi-layer perceptron (MLP)), as a function of the number of observed points and signal to noise ratio (SNR). Default parameters: SNR = 2.5, $N=100$ observed points. 
  \textbf{Right:} Classification accuracy of SPE, averaged over 1000 randomly sampled datasets (see \Cref{app:sparse-classif-eval}), across different dynamical systems (simple oscillator (SO), BZ reaction, Sel\'kov, Li\'enard sigmoid, Li\'enard polynomial and Van der Pol).
  Average accuracy over all systems is shown in black. Default parameters: SNR = 100, $N=500$ observed points.
  In all plots, shaded areas correspond to standard error of the mean.
  }
  \label{fig:sparse-results}  
\end{figure}

To quantitatively evaluate the performance of SPE, we calculate the Wasserstein ($W_2$) distance between samples from the invariant set of the ground-truth system and those from the invariant set of the fitted model (details in \Cref{app:invariant-eval}). 
When the positions $x_i$ are sparse, or the vectors are noisy, SPE more accurately reconstructs the underlying invariant sets than existing baselines, including a $k$\textbf{-nearest neighbor} (kNN) interpolator, \textbf{SINDy} \citep{brunton2016sindy}, and a multilayer perceptron (\textbf{MLP}) (\Cref{fig:sparse-results} left; baseline details are in \Cref{app:inv-baselines}). In particular, when the sparsity level or noise level are high, SPE outperforms existing baselines more substantially (\Cref{fig:sparse-results}, left), as it is constrained to model only stable dynamics and can therefore more closely reconstruct the underlying vector field. 

\begin{figure*}[!tbh]
\begin{center}
    \includegraphics[width=\linewidth]{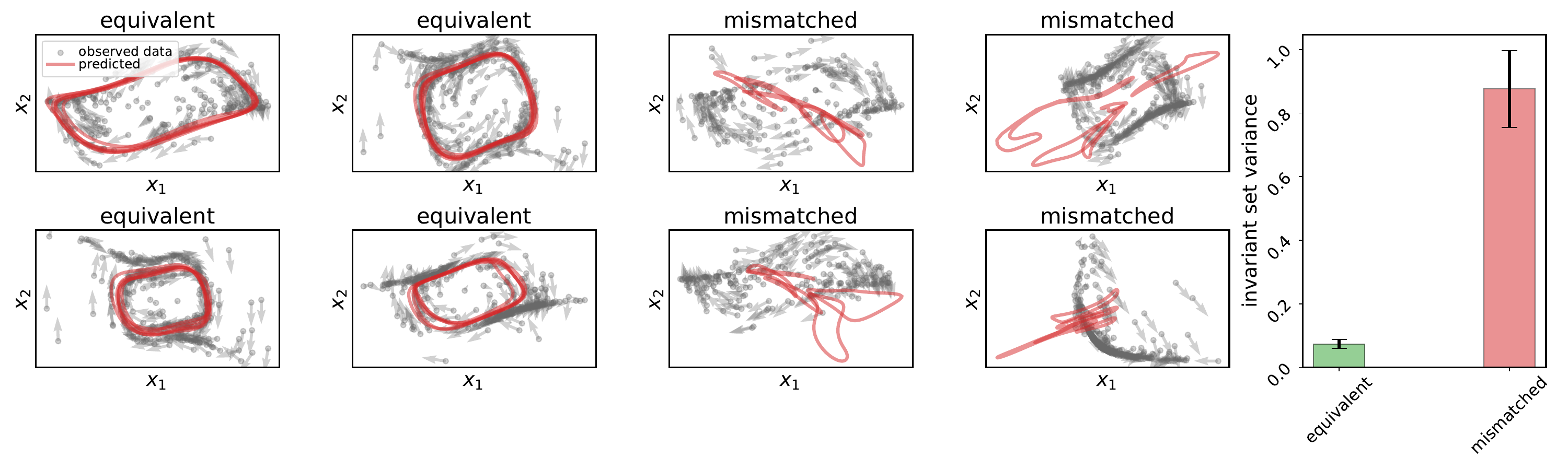}
\end{center}
  \caption{\textbf{Detecting prototype mismatch through variance of the estimated invariant set.} 
  \textbf{Left:} Shown are represented examples for observed vectors from systems exhibiting limit cycles (gray, from the simple oscillator, BZ reaction, Sel\'kov, Li\'enard sigmoid, Li\'enard polynomial and Van der Pol families of systems \Cref{app:2D-systems}), and predicted invariant sets by SPE (red) based on oscillatory prototypes with either clockwise or counter-clockwise angular motion (\Cref{app:mismatch-details}). Data and prototype are considered equivalent when they share the same direction of angular velocity, or mismatched when they do not.
  When the two systems are equivalent, the location of the estimated invariant set is robust to subsampling of the data (where randomly-chosen 50\% of the data points were omitted) (left two columns). Conversely, when there is a mismatch between the prototype and the system underlying the observed data, the learned diffeomorphism tends to overfit, and different random subsampling of the data yields substantially different estimated limit cycles (red) (right two columns). \textbf{Right:} The variance in the estimated positions of the invariant set, averaged over the whole invariant set, is significantly higher for mismatched relative to equivalent prototype-system pairs ($p < 0.05$). The error bars represent the standard error of the mean of the variance.}
  \label{fig:invariant-set-variance}
\end{figure*}

SPE can avoid overfitting by detecting mismatches between input prototypes and the data through sensitivity analysis, where small changes in the data will result in substantially different fitted diffeomorphism for mismatched prototypes.
Indeed, we observe across 300 systems-prototypes pairs that fitted invariant sets by SPE for matched prototypes (reflecting the underlying dynamics in the data) are robust to random splits of the data and exhibit low variance of invariant set localization, relative to mismatched prototype-system pairs (\Cref{fig:invariant-set-variance}, more details in \Cref{app:mismatch-details}).
Thus, this measure of fitting variance, paired with the equivalence loss, can be used to detect out-of-dictionary scenarios, where the prototypes and data do not exhibit the same behavior.

\subsection{Classifying dynamical systems with prototype equivalences}\label{sec:classify}

SPE can extend beyond the characterization of a given invariant set based on a single prototype, to distinguishing between different potential prototypes using the relative loss in fitting multiple prototypes.
As our prototype set, we choose all four effective behaviors implied by \Cref{eq:prot}, with $a=\pm\frac{1}{4}$ and $\omega=\pm\frac{1}{2}$. 
Four diffeomorphisms, $H_k(x)$, were fit to each observed system and the one with the minimal value of the equivalence loss defined in \Cref{eq:diffeo-loss} was used to determine class membership. 

To test the capabilities of SPE in sparse and noisy scenarios, in \Cref{fig:sparse-results} (right) we track the classification accuracy when decreasing the number of observed vectors $N$ and under increasingly noisy scenarios, with low signal-to-noise ratios (SNRs) (see \Cref{app:sparse-classif-eval} for details regarding data). The decline in performance is gradual along both of these axes - even when classifying using only 50 random vectors, average accuracy was still relatively high, at $\sim70\%$. 
Beyond these results, in \Cref{app:classif-baselines} we also compared the performance of SPE on the task of classification in the low-noise regime to baselines which take as input a dense grid of the dynamics. We found that SPE outperforms all baselines on average accuracy in the classification of the considered family of systems. 

SPE's classification and localization performance is not constrained to limit cycles and node attractors, and can be flexibly generalized to additional prototypes. To demonstrate such capabilities, we evaluated the performance of SPE on a challenging classification of multistable systems with a varying number of basins of attraction (\Cref{app:multistable}). Our results show that SPE can successfully distinguish between dynamical systems with varying numbers of fixed points, and can additionally localize them successfully (\Cref{fig:multistable}). Furthermore, as the loss used to train SPE corresponds to a graded notion of equivalence, we found that SPE is also able to capture dominant behaviors of the system even when there is a mismatch between the true behavior and the chosen prototype (\Cref{app:mismatch}).

\subsection{Localizing and classifying invariant sets in the synthetic repressilator system}\label{sec:highdim-prototypes}
\begin{figure*}[!tbh]
\begin{center}
    \includegraphics[width=\linewidth]{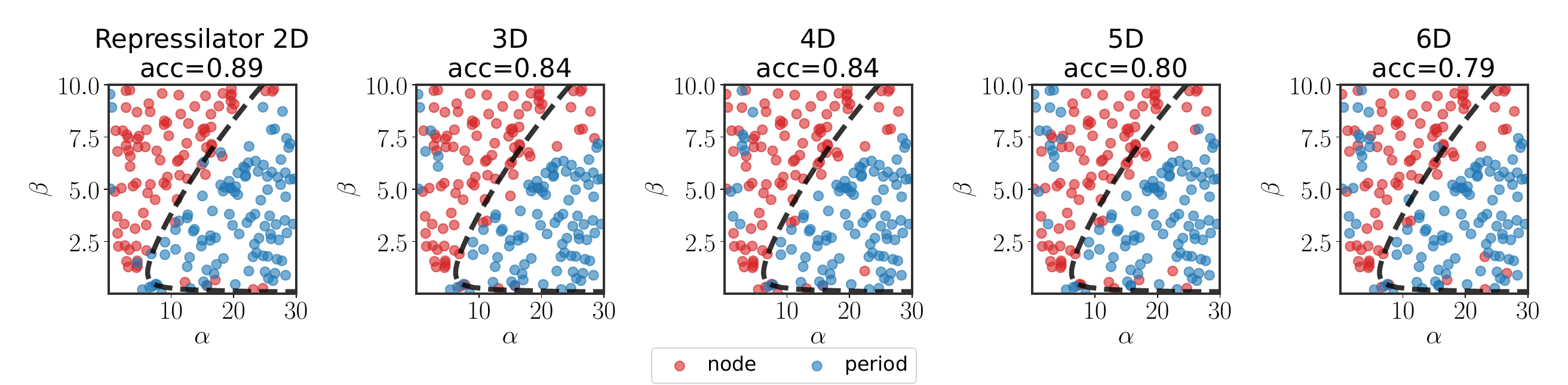}
    \includegraphics[width=\linewidth]{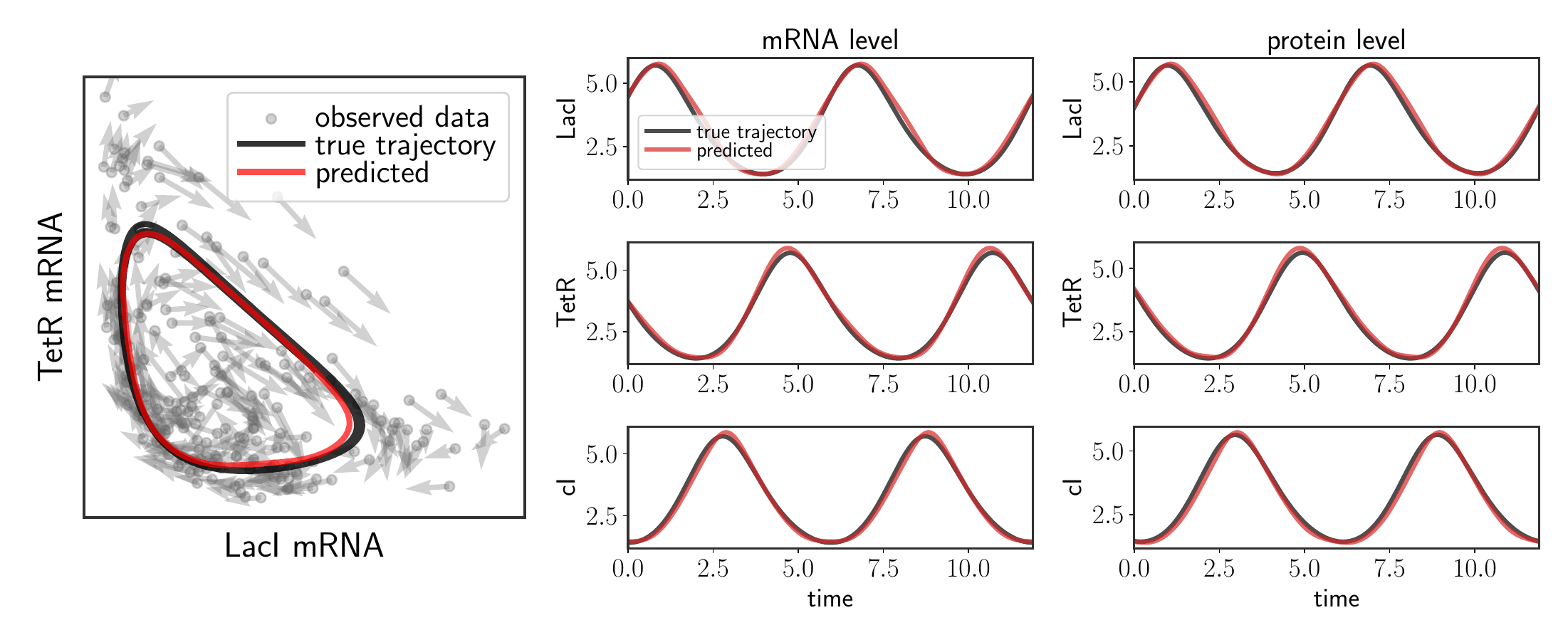}
\end{center}
  \caption{ \textbf{Characterizing dynamics in higher dimensions and recovering the invariant structure of the repressilator genetic circuit.}
  \textbf{Top:} 
  In each scatter plot, each point represents a repressilator system with different order parameters chosen to emulate a bifurcation between periodic and node behavior (full details in \Cref{app:repressilator-sim}). Each system was classified by SPE as either a 2D embedded limit cycle (blue) or node attractor (red). The dashed black line depicts the ground truth bifurcation boundary between periodic and node behavior. In each column, only a subset of dimensions were used for the classification, from 2D (left, LacI-TetR mRNA plane) to the full 6D (right).
  \textbf{Bottom:} 
  An example of the invariant set predicted by SPE (red) from the observed vectors (gray), projected from the 6D space to the 2D LacI-Tetr mRNA plane for visualization. A ground-truth trajectory was simulated (black) for visual comparison.
  Time series of the mRNA (center) and protein (right) expression levels can be extracted from the predicted limit cycle (red). These are overlayed on top of trajectories simulated from the underlying (hidden) system (black).
  }
  \label{fig:repressilator}
\end{figure*}

To test our approach in higher-dimensional systems, we first classify and localize invariant structures in a synthetic repressilator model of gene regulation \citep{elowitz2000repressilator} (see \Cref{app:repressilator-sim} for governing equations). The repressilator models a genetic regulatory network in bacteria, based on the mRNA and protein counts of three genes - TetR, LacI and $\lambda$ phage cI - which inhibit each other in a cyclic fashion. 
The repressilator system acts as a simplified model for clock circuits in biological systems, and was one of the first oscillating regulatory circuits that was synthetically engineered  \citep{elowitz2000repressilator}.
The repressilator system spans six dimensions, but under specific parameter values it exhibits either a point attractor or contains an embedded 2D limit cycle \citep{Verdugo2018-np,Potvin-Trottier2016-wj}. 
To model simulated data from this system, we use higher-dimensional prototypes that are the combination of two behaviors: (1) a 2D limit cycle/equilibrium behavior as in \Cref{eq:prot}; (2) exponential decay to a fixed point along variables uncoupled from the embedded 2D system. Mathematically, we define these prototypes as:
\begin{equation} \label{eq:multidim}
    g(y)=\left[g_\text{2D}(y_1),\quad -\tau\cdot y_2\right]^T
\end{equation}
with $y_1\in\mathbb{R}^2$ representing the first 2 dimensions of $y$ and $y_2\in\mathbb{R}^{d-2}$ the remaining dimensions. $g_\text{2D}(\cdot)$ are velocities in two dimensions, and $\tau$ is a decay factor. We used $\tau=1/2$ in all of our experiments.

We then classify repressilator systems into node or cycle attractors using SPE, when the systems vary in terms of transcription $\alpha$, and protein/mRNA degradation rates $\beta$. For each system, we simulate $N=1000$ observed pairs of positions and vectors using the same sampling scheme for the data as described in \Cref{sec:recon}, \Cref{fig:repressilator}. To compare the performance along different dimensionalities, we project the observed positions $x_i$ and velocities $\dot{x}_i$ onto a predefined subset of the six dimensions (see \Cref{app:repressilator-sim} for details). We observed (\Cref{fig:repressilator}) that our model can classify two-dimensional repressilator systems better than earlier work \citep{moriel2023let}, whose accuracy was 87\% (next to our 89\%), and that SPE can accurately distinguish point from cyclic behavior across all dimensionalities considered. 
Classification of this kind, carried out explicitly in higher dimensionalities, is infeasible using techniques from previous work. Moreover, our optimized diffeomorphisms in the six-dimensional case can be used to accurately locate the cyclic dynamics of the repressilator system, recapitulating the behavior of the system and its invariant set (\Cref{fig:repressilator}). 

\subsection{Uncovering cell cycle expression patterns in single-cell data}\label{sec:fucci}
\begin{figure*}[!tbh]
\begin{center}
    \includegraphics[width=\linewidth]{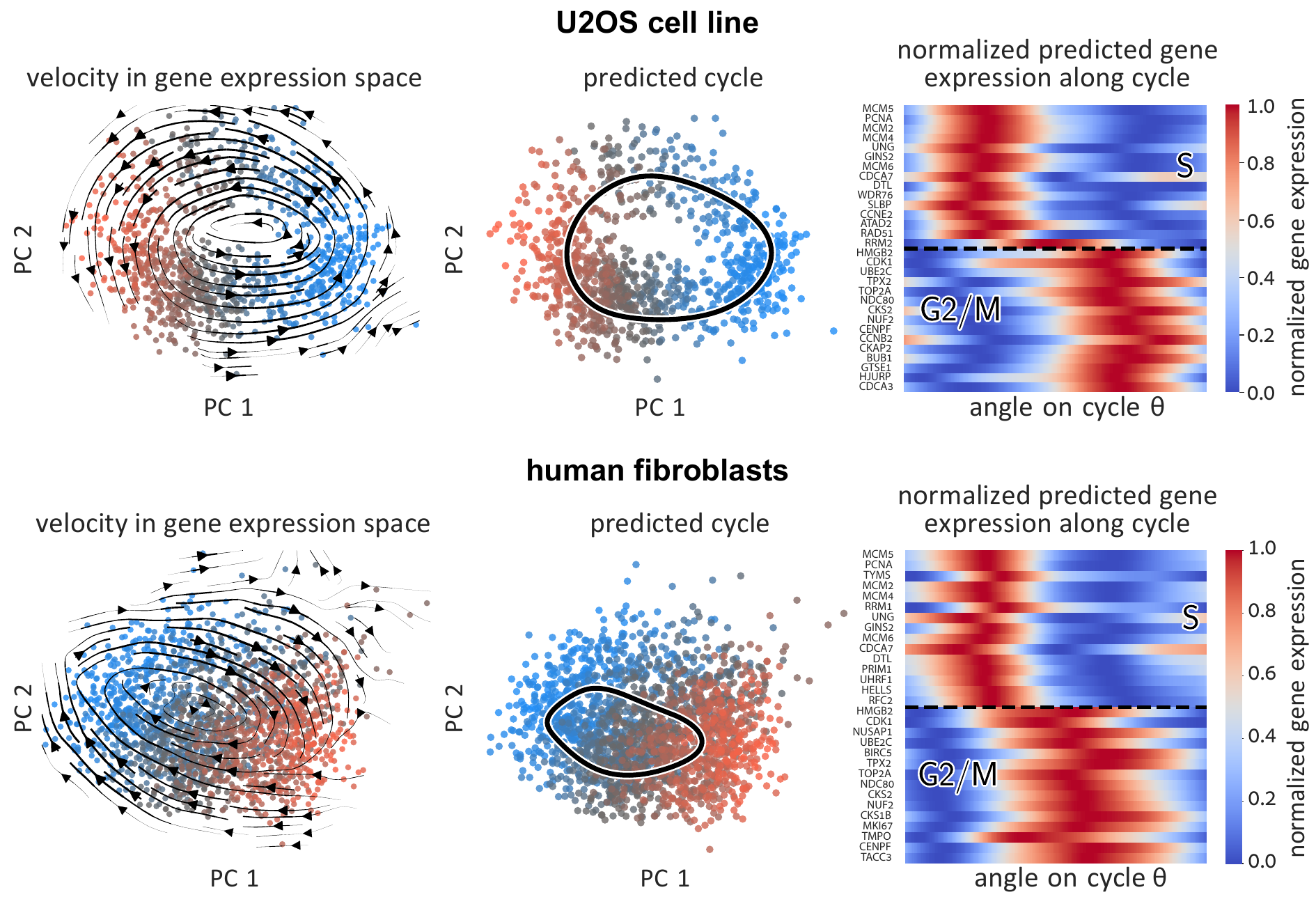}
\end{center}
  \caption{\textbf{Recovering the oscillatory gene expression patterns of the cell cycle from single-cell RNA-sequencing data.} Results are shown for scRNA-seq of U2OS cell line (top) \citet{mahdessian2021fuccicycle} and human fibroblast cells (bottom) \citet{riba2022cycle}.
  \textbf{Left:} Gene expression RNA velocity of cells undergoing proliferation in 2D PCA space, estimated using scVelo \cite{bergen2020scvelo}. Colors denote the cycle phase score based on marker gene expression; red corresponds to high expression of S-phase markers, blue to high expression of G2/M-phase genes, and gray to low levels of both (details in \Cref{app:fucci}). 
  \textbf{Middle:} SPE was used to fit a periodic prototype from \Cref{sec:highdim-prototypes} to PCA projections of the RNA velocity data (100 dimensions for U2OS data, 10 dimensions for human fibroblast data); the resulting attractor (black), along with the sampled cells (locations and colors correspond to left column), were projected onto 2D PCA space. \textbf{Right:} Normalized gene expressions along the cycle attractor predicted by SPE, as a function of the angle along the limit cycle ($\theta$), for a set of marker genes of S phase (top) and G2/M phase (bottom). See details in \Cref{app:fucci}.
  }
  \label{fig:fucci}
\end{figure*}

Finally, we show how SPE can be used to trace the periodic gene expression dynamics of proliferating cells. This demonstrates our method’s ability to recover biologically meaningful periodic processes directly from high-dimensional, noisy and sparse experimental data.
Here, we analyze single-cell RNA-sequencing (scRNA-seq) data from two human cell lines, profiling over thousands of genes across thousands of cells \citep{mahdessian2021fuccicycle,riba2022cycle}. 
Previous approaches have inferred cell-cycle progression using marker genes \citep{tirosh2016dissecting}, a learned projection \citep{Zheng2022-iw,riba2022cycle}, or using structural priors \citep{Schwabe2020-xh,karin2023scprisma}. However, these existing approaches either do not allow for simulation of future states as they do not directly model the dynamics, or rely on explicit knowledge of cell cycle marker genes, information that is not necessarily available for other processes.

For each scRNA-seq dataset, we infer local velocities of cells in gene expression space using scVelo \citep{bergen2020scvelo}, which computes high-dimensional RNA velocity vectors for each of the sampled cells. 
We reduce the dimensionality of the gene expression profiles and the respective velocities using principal component analysis (PCA) to 10-100 dimensions (\Cref{fig:fucci}), enough to capture the variability in the data. This projection reduces the computational cost of the fitting procedure, while providing an approximate map back from the lower-dimensional prototype space to the full gene expression space.
We then fit a high-dimensional limit-cycle attractor (as described in \Cref{eq:multidim}) to the RNA velocity data, capturing the periodic nature of the cell cycle (\Cref{fig:fucci}, middle, see \Cref{app:fucci} for more details on data and prototype). 

The expression of cell-cycle genes follows a well-defined trajectory through different phases: initial growth (G1), DNA replication (S phase), further growth (G2), and mitosis (M). 
Utilizing the PCA projection, we can map the learned attractor back to gene expression space, revealing periodic, anti-correlated expression patterns of key cell-cycle marker genes for the S and G2/M-phase (\Cref{fig:fucci}, right). 

\section{Discussion}
Our proposed approach, \emph{smooth prototype equivalences} (SPE), addresses a challenging problem in data-driven dynamical systems: characterizing long-term behavior from sparse, noisy data. 
The key idea is to learn the correspondence between idealized prototype dynamics and their noisy, real-world counterparts in a way that reveals long-term structure. To do so, an invertible neural network is trained to capture the mapping from data space to prototype space. 
We demonstrated the efficacy and efficiency of this approach in identifying and tracing invariant set structures and classifying between multiple potential long-term behaviors  on both challenging synthetic and real-world data sets.  

In this work, our focus was the detection and classification of long-term behaviors that are robust to noise, such as limit cycles and node attractors. 
Indeed, our main focus were phenomena from empirical biological systems, where dynamics reflecting and driving collective cellular functions are robust to perturbations, including differentiation and oscillatory processes such as the circadian rhythm and the cell cycle. 
At the other side of this scale are systems which are structurally unstable and include chaotic systems with strange attractors. Extending prototype equivalence approaches to such systems, whose underlying vector fields are sensitive to small perturbations, and where identifying smoothly equivalent prototypes could be a challenging task by itself, is an important future challenge. 

While SPE relies on a set of pre-defined prototypes, it is a useful tool for characterizing dynamical systems from realistic data due to its relaxed notion of equivalence, where an optimized transformation between prototype and data can be found, so that even a partially matching prototype can be useful for the localization and categorization of dynamical behaviors (\Cref{app:mismatch}). 

To parameterize our diffeomorphism, we used invertible neural networks (INNs) based on affine coupling layers. To ensure that our INNs are expressive while remaining lightweight and easily differentiable, we implemented a specific form of coupling which we termed Fourier feature coupling.
Other options for parameteric invertible mappings exist, such as invertible ResNets \citep{behrmann2019invertible, chen2024dform} or Neural ODEs \citep{chen2018neural}. However, while these options are typically more expressive than INNs adapted from normalizing flows (NFs), they come with steep overheads in the calculations of the inverse function, Jacobian-vector products (JVPs) and their log-determinants. Therefore, we opted for sufficiently expressive networks where needed components (inverse, JVPs, log-determinants) could be easily calculated.

While in this work our primary focus were locally simple behaviors, such as node or cycle attractors or a small collection of node attractors (\Cref{app:multistable}), an important future step would entail the combination of multiple local dynamics into a larger, more complex system.
This could be through the use of mixtures of experts \citep{jacobs1991adaptive}, the incorporation of more complicated prototypes, or the combination of SPE together with other methodologies from the field of data-driven nonlinear dynamics. In all cases, SPE could be potentially used as an interpretable baseline component, capturing intricate dynamics in many dimensions.
We also envisage that SPE can be paired with more advanced parameterizations of INNs in order to tackle more complex behaviors, in higher dimensional settings, which may take inspiration from computer vision \citep{papamakarios2021normalizing,dinh2014nice,durkan2019splineflow}. 

Our method directly addresses challenges faced in scientific domains such as biology, neuroscience, and physics. By offering an equation-free approach for the detection and classification of invariant sets, our method has the potential to advance the identification and interpretation of complex, nonlinear processes given sparse, noisy, high-dimensional experimental measurements.

\section*{Data availability}
All cell cycle data from \Cref{sec:fucci} are publicly available \citep{mahdessian2021fuccicycle,riba2022cycle}.
The code is publicly available at: \href{https://github.com/nitzanlab/prototype-equivalences}{https://github.com/nitzanlab/prototype-equivalences}.




\appendix
\section{SPE method}\label{app:diff-details}

The diffeomorphisms $H:\mathcal{X}\rightarrow\mathcal{Y}$ used for SPE are invertible neural networks (INNs) with alternating \emph{Affine} and \emph{Fourier Feature Coupling} (FFCoupling) transforms, both of which are explained in further details below. Furthermore, the first layer of the normalizing flow was set to standardize the data, effectively an \emph{ActNorm} layer~\cite{papamakarios2021normalizing}, defined as:
\begin{equation}
    [\text{ActNorm(x)}]_i = \frac{x_i-\text{mean}(x_i)}{\text{std}(x_i)}
\end{equation}
In the above, $\text{mean}(x_i)$ and $\text{std}(x_i)$ are the mean and standard deviations of the $i$-th coordinate (e.g. the $i$-th gene) of the observed data, respectively. This ActNorm layer was kept frozen after initialization.

\subsection{Fourier feature coupling}\label{app:ffcoupling}

The FFCoupling layer we used is a modification of the standard \emph{AffineCoupling}~\cite{papamakarios2021normalizing,dinh2014nice} layer used in normalizing flows. AffineCoupling splits an input $x$ to two sub-vectors $x_1$ and $x_2$, and the transformation itself is defined as:
\begin{equation}
    \left[\begin{matrix}
        y_1\\y_2
    \end{matrix}\right]=\text{AffineCoupling}(x)=\left[\begin{matrix}
        x_1\\\exp[f_s(x_1)]\circ x_2 + f_t(x_1)
    \end{matrix}\right]
\end{equation}
where $f_s(\cdot)$ is a scaling function, $\exp[f_s(\cdot)]$ is the element-wise exponent of $f_s(\cdot)$, $f_t(\cdot)$ is a translation (or displacement) function, and $\circ$ stands for element-wise multiplication. AffineCoupling is defined in this manner to ensure invertibility, in which case $f_s(\cdot)$ and $f_t(\cdot)$ can be arbitrarily complex functions. In previous applications \cite{dinh2014nice}, these two functions are typically parameterized by a multi-layer perceptron (MLP). Baked into the AffineCoupling transformation is the simple form of its log-determinant:
\begin{equation}
    \log \left|\text{AffineCoupling}(x)\right|=\sum_i [f_s(x_1)]_i
\end{equation}
Since AffineCoupling only transforms $x_2$ as a function of $x_1$, it is standard to chain two transformations, one where $x_1$ acts on $x_2$ and another with the roles reversed. For simplicity, we regard such a chaining of two AffineCoupling transformations as a single layer.

When both the scale and the translation functions are defined as MLPs, calculating the Jacobian or Jacobian-vector products (JVPs) requires autograd operations, and cannot be calculated in closed form.
These JVPs need to be computed for SPE's loss (\Cref{eq:diffeo-loss}) and can potentially be found using autograd methods, albeit at a high computational cost. 
Instead, we opt to use transformations whose JVPs can be calculated in closed-form, while remaining expressive. 
Therefore, for our scale and translation functions we use the following \emph{FFCoupling}:
\begin{equation}
    \mathcal{F}(x) = \sum_{k=1}^K\Theta_k\cos\left(\frac{2\pi}{R}k\cdot x+\phi_k\right)
\end{equation}
where $\cos(x)$ is an element-wise cosine. As defined, the frequency coefficients $\Theta_k\in\mathbb{R}^{d\times d}$ and phases $\phi_k\in\mathbb{R}^{d}$ are learnable parameters, while the number of frequencies $K$ and the range $R$ are hyperparameters. This transformation is a parametric form of a discrete cosine transform, such that when $K$ is large it can approximate any smooth function bounded in the range $[-R/2,R/2]$. It also admits a very simple form for the Jacobian, given by:
\begin{equation}
    \frac{\partial}{\partial x}\mathcal{F}(x)=-\frac{2\pi}{R}\sum_k k\cdot \Theta_k\ \text{diag}\left[\sin\left(\frac{2\pi}{R}k\cdot x+\phi_{k}\right)\right]
\end{equation}

\subsection{Affine transformation layer}
In addition to the FFCoupling layers, we also use affine transformations. To ensure invertibility, we parameterize these transformations as:
\begin{equation}
    \text{Affine}(x)=WW^Tx + e^{\varphi}\cdot x + \mu
\end{equation}
where $W\in\mathbb{R}^{d\times q}$ is a low-rank weight matrix with rank $q$, $\varphi\in\mathbb{R}$ is a scalar added to the diagonal of the low rank matrix $W^TW$ to ensure that the whole transformation is invertible, and $\mu\in\mathbb{R}^d$ represents a translation.
$W$, $\varphi$ and $\mu$ are all learnable parameters.
This factorization ensures that the matrix $WW^T+Ie^{\varphi}$ is positive-definite, and thus invertible. It also enables efficient inversion using the matrix-inversion lemma:
\begin{equation}
    \left(WW^T+Ie^\varphi\right)^{-1}=Ie^{-\varphi}-W\left(W^TW+Ie^{\varphi}\right)^{-1}W^T
\end{equation}
which involves inverting a $q\times q$ matrix ($W^TW$), instead of a $d\times d$ matrix ($WW^T$).

\subsection{Householder transformations}

In addition to the Affine and FFCoupling layers, for higher dimension data we also used \emph{Householder transformations} \citep{papamakarios2021normalizing,tomczak2016improving}, which act as building blocks for orthogonal transformations. The Householder transformation is defined as:
\begin{equation}
    \text{HH}(x)=\left(I-\frac{2}{\|v\|^2}vv^T\right)x
\end{equation}
where $v\in\mathbb{R}^d$ are the parameters of the transformation. This transformation is orthogonal, and so has a determinant of 1. Furthermore, it can be shown \citep{tomczak2016improving} that any rotation in a $d$-subspace can be described by $d-1$ of these transformations. Thus, these Householder layers can directly rotate space, unlike the Affine transformations defined above which are more constrained.

\subsection{Network specifications}\label{app:train-info}

As mentioned above, the INNs we use have alternating Affine and FFCoupling transforms. A single block of our network is defined as Affine $\rightarrow$ FFCoupling $\rightarrow$ ReverseFFCoupling, where the ReverseFFCoupling switches the roles of $x_1$ and $x_2$, as explained in \Cref{app:ffcoupling}. 

Before these blocks, our networks include a frozen ActNorm layer that standardizes the data and ensures that all learned diffeomorphisms have similar inputs, and then a full-rank Affine transformation, with learnable weights. For the experiments performed in \Cref{sec:fucci} and \Cref{app:multistable} we added 2 Householder transforms after the frozen ActNorm layer and before the Affine and FFCoupling layers.

\subsection{Projection regularization}\label{app:projection}

For high-dimensional prototypes (\Cref{eq:multidim}), 
the diffeomorphisms are regularized to map the data as close as possible to the invariant sets of the embedded 2D dynamics, in accordance with the prototypes construction. 
Specifically, the observed points are projected onto the 2D invariant set of the prototype using the diffeomorphism, and then back into data space as follows:
\begin{equation}
    P(x;H_\theta)=H_\theta^{-1}\left([H_\theta(x)]_1, [H_\theta(x)]_2,0,\cdots,0\right)
\end{equation}
where $[H_\theta(x)]_i$ is the $i$-th coordinate of $H_\theta(x)$ and $P(x;H_\theta)$ denotes the projection according to $H_\theta(x)$. 

The projection into data space is modeled by a Gaussian observation model, with the following loss added when training:
\begin{equation}
    L_\text{proj}(H_\theta,g,\lambda_\text{proj})=e^{\lambda_\text{proj}}\frac{1}{N}\sum_i^N\left\|x_i - P(x_i;H_\theta)\right\|^2 - \frac{1}{N}\lambda_\text{proj}
\end{equation}\label{eq:proj-loss}
where $\lambda_\text{proj}$ is a regularization coefficient, which takes the form of the log of the precision of the Gaussian observation model. Adding the regularization coefficient in this manner means that it can be optimized in parallel to the diffeomorphism, and allows for an adaptive fitting process. When modeled in this way, the projection is similar to generative topographic mapping (GTM) \cite{bishop1998gtm} when the matching between the latent codes and the data is known. 

\subsection{Optimization criteria}\label{app:opt-criteria}

The main term in the optimization objective of our diffeomorphisms is the equivalence loss (\Cref{eq:diffeo-loss}), given by:
\begin{equation}
    L_\text{E}(H_\theta, g) = \frac{1}{N}\sum_{i=1}^N \left\|\frac{\partial_{x_i} H_\theta(x_i)\dot{x}_i}{\|\partial_{x_i} H_\theta(x_i)\dot{x}_i\|}-\frac{g\left(H_\theta(x_i)\right)}{\| g\left(H_\theta(x_i)\right)\|}\right\|^2 
\end{equation}
In addition to the above loss, we use several regularizations to encourage convergence, such as the projection regularization described in \Cref{app:projection} and weight decay.
We also penalize networks excessively expanding or contracting space by regularizing the absolute value of the log-determinant to be close to 0: 
\begin{equation}
\text{L}_{\text{det}}(H_\theta)=\frac{1}{N}\sum_{i=1}^N\left|\log \text{det}\left[\partial_{x_i}H_\theta(x_i)\right]\right|
\end{equation}
The full loss used to optimize the diffeomorphism $H$ in SPE is thus given by:
\begin{equation}
    L_\text{tot}(H_\theta,g)=L_\text{E}(H_\theta,g)+L_\text{proj}(H_\theta,g,\lambda_\text{proj}) + \lambda_\text{det}\text{L}_{\text{det}}(H_\theta).
\end{equation}\label{eq:full-loss}
where $\lambda_\text{det}>0$ is a regularization coefficient.

All training is carried out using the Adam optimizer \cite{kingma2014adam} with weight decay. 


\subsection{Hyperparameters settings}\label{app:hyperparams}

For all simulated systems, hyperparameters were chosen to maximize performance on a separate, held out validation set of 100 datasets which were not used for evaluation. 
In all experiments, unless specified otherwise, we used the simple oscillator (SO) prototype defined in \Cref{eq:prot}. The hyperparameters of the prototype are $a$, the radius of the limit cycle, and $\omega$, the angular velocity.

\begin{table}[htb!]
  \centering
  \small
  \caption{\textbf{Hyperparameter settings} }
  \resizebox{\textwidth}{!}{
    \begin{tabular}{|r|c|c|c|c|c|c|c|c|c|}
      \hline
        Experiment & \# blocks & $K$ & iterations & learning rate & weight decay & $\lambda_\text{det}$ & $\lambda_\text{proj}$ & prototype $a/\omega$ \\
      \hline
      \Cref{fig:invariants,fig:sparse-results,fig:invariant-set-variance,fig:mismatch,fig:more-examples} & 4 & 5 & 500 & $10^{-3}$ & $10^{-3}$ & $10^{-3}$ & - & $\pm0.25/\pm0.5$ \\
      \Cref{fig:repressilator} & 3 & 5 & 600 & $10^{-3}$ & $10^{-3}$ & $10^{-3}$ & $e^{-1}$ & $\pm0.25/\pm2$ \\
      \Cref{fig:fucci} & 3 & 3 & 1,000 & $5\cdot10^{-4}$ & $10^{-3}$ & 0 & $e^{-2}$ & $0.1/5$ \\
      \Cref{fig:grid-results} & 3 & 5 & 600 & $10^{-3}$ & 0 & 0 & - & $\pm0.25/\pm0.5$ \\
      \hline
    \end{tabular}
  }
\label{tab:hyperparameters}
\end{table}

As our INNs are relatively small, and most of the data is of a low dimension (relative to typical machine learning tasks), a single CPU with 16GB RAM was sufficient for training our networks. Fitting a single INN takes less than 10 seconds for all of the 2D simulated data and $\sim$90-120 seconds for the cell-cycle data.

\subsection{Effects of hyperparameters on performance}

\begin{figure*}[!tbh]
\begin{center}
    \includegraphics[width=.6\linewidth]{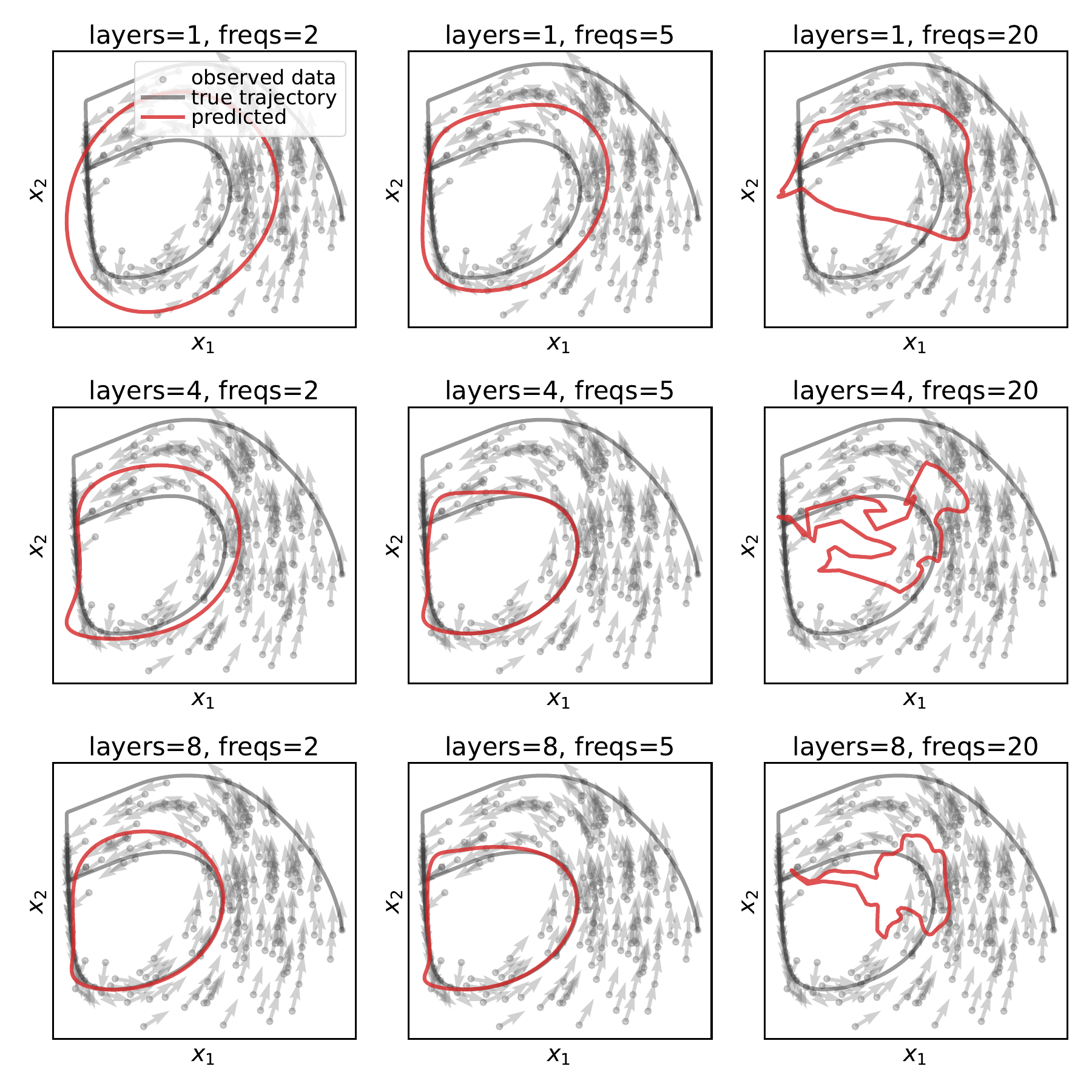}
\end{center}
  \caption{\textbf{Estimation of limit cycles using different numbers of layers and frequencies in the diffeomorphism $H_\theta$.}
    Examples of invariant sets predicted by SPE (red) from observed vectors (gray), for a BZ reaction system (\Cref{app:2D-systems}) exhibiting a limit cycle attractor. Ground-truth trajectories were simulated for each hidden system (black) for visual comparison.
    }
  \label{fig:effects-layers-freqs}
\end{figure*}
\begin{figure*}[!tbh]
\begin{center}
    \includegraphics[width=.8\linewidth]{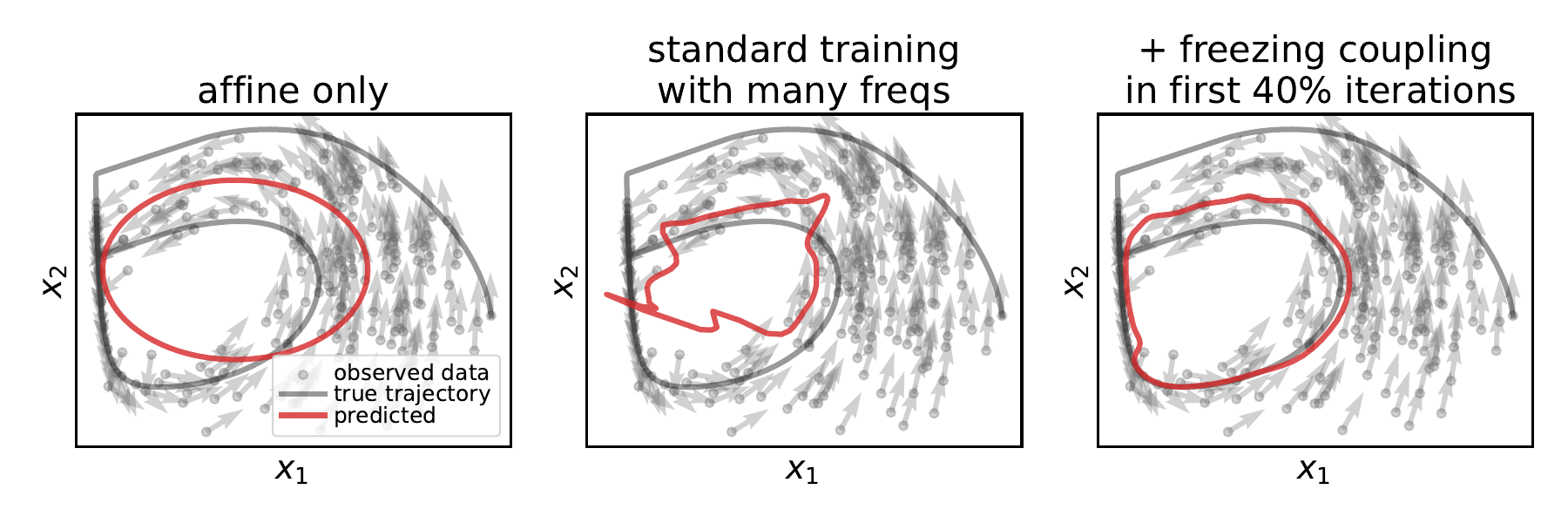}
\end{center}
  \caption{\textbf{Initializing network from the solution of an affine-only diffeomorphism improves fitting.} \textbf{Left}: when the invariant set is not an ellipse, using a diffeomorphism defined only by an affine transformation does not faithfully capture the underlying invariant set. \textbf{Middle:} if too many frequencies are used with SPE in such a case, the model overfits. \textbf{Right:} when freezing all coupling layers for the initial iterations of the fitting procedure, SPE converges more faithfully to the true invariant set.
  }
  \label{fig:more-freqs}
\end{figure*}

The classification results by SPE were robust to the choice of number of frequencies and layers, as demonstrated in the table below. 

\begin{table}[htb!]
  \centering
  \small
  \caption{\textbf{Classification performance for different numbers of layers and frequencies of the trained invertible neural network.} Numbers correspond to the average accuracy over 1000 different systems uniformly sampled from the set of systems in \Cref{tab:gov-eqs}, with $N=250$ observations per system.}\label{tab:ablation}
  \resizebox{.5\textwidth}{!}{
    \begin{tabular}{|r|cccccc|}
      \hline
        frequencies ($K$) & 2& 3 & 4 & 5 & 6 & 7 \\
      \hline
      2 layers & 88.7\% & 84.9\% & 89.1\% & 85.8\% & 84.7\% & 81.3\% \\
      3 layers & 84.9\% & 86.1\% & 84.9\% & 83.6\% & 83.2\% & 80.1\% \\
      4 layers & 84\% & 86.8\% & 83.5\% & 81.5\% & 79.7\% & 80.9\% \\
      5 layers & 82.2\% & 84.9\% & 82.5\% & 80.3\% & 77.8\% & 77.3\% \\
      \hline
    \end{tabular}
  }
\label{tab:classif_ablations}
\end{table}

In our experiments, we remained within the robust range demonstrated in \Cref{tab:ablation}, avoiding too many frequencies that can lead to overfit to overly complex estimations of the invariant set
(\Cref{fig:effects-layers-freqs}), while also avoiding too few frequencies and layers to keep the network expressive enough to capture non-elliptical invariant sets. 

If many frequencies are needed, for instance when the invariant set has a complex structure, we found it beneficial to begin training with frozen coupling layers, as shown in \Cref{fig:more-freqs}. This essentially initializes the network at the solution found when the diffeomorphism is constrained to be affine, after which the behavior is fine-tuned using the available frequencies.

\section{Simulation details}\label{app:sim-info}

\subsection{2D systems}\label{app:2D-systems}
\begin{figure*}[!tbh]
\begin{center}
    \includegraphics[width=\linewidth]{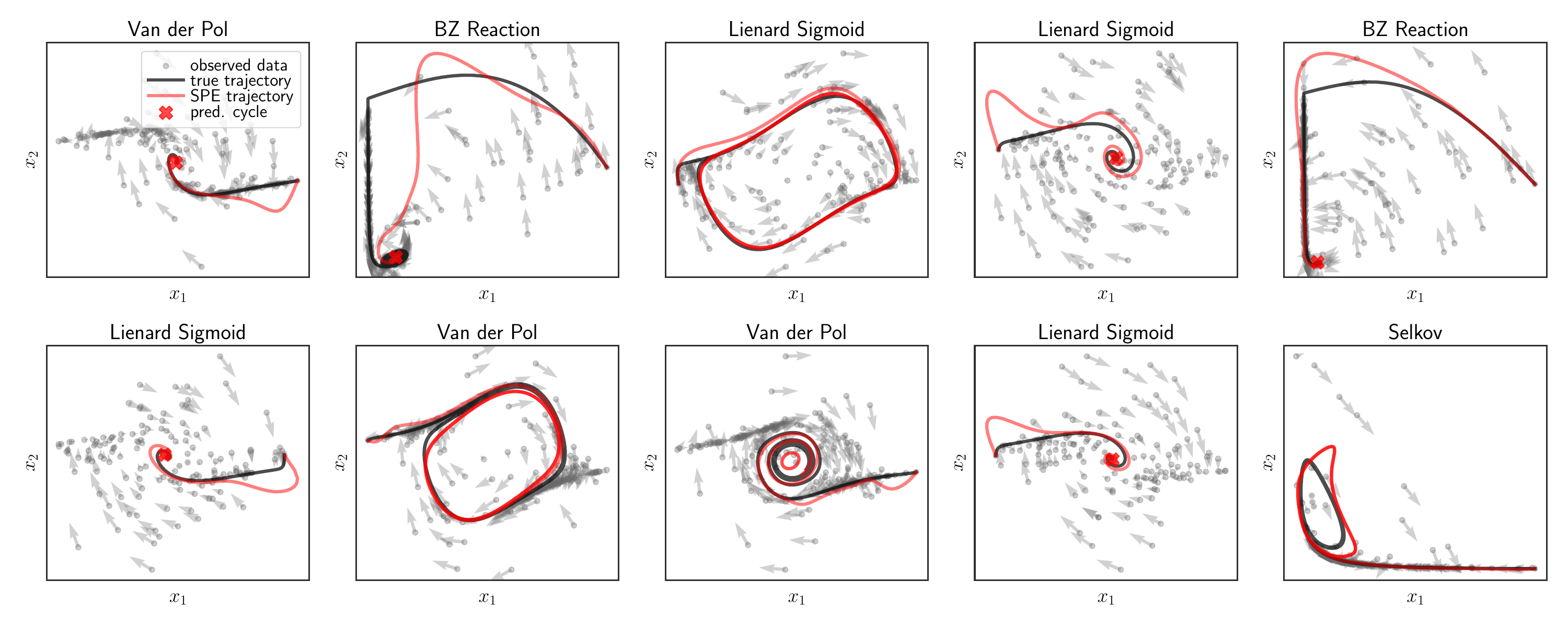}
\end{center}
  \caption{\textbf{Randomly picked examples to localizing invariant sets using SPE on synthetic data from \Cref{app:2D-systems}.}
  In all of the above, solid black lines depict ground-truth trajectories simulated according to the true governing equations which were used to generate the data, represented by the gray vectors. The solid red lines are trajectories predicted by SPE. 
   SPE localizes the invariant set from the data, even when transient states are ambiguous in the observed data.
  }
  \label{fig:more-examples}
\end{figure*}
For the 2D simulated dynamics (\Cref{sec:recon}), we use the same systems as in \citet{moriel2023let}, whose governing equations appear in  \Cref{tab:gov-eqs}. 
\begin{table}[htb!]
  \small
  \caption{Simulated dynamical systems. Columns correspond to dynamical system type, corresponding governing equations, range of phase space used for initial conditions, and range of parameter values used in our simulations, and parameter settings that correspond to a limit cycle. 
  }
  \resizebox{\textwidth}{!}{
    \begin{tabular}{|c|c|c|c|c|}
      \hline
      \textbf{System Name} & \textbf{Equation} & \textbf{Initial cond.} & \textbf{Parameter ranges}  & \textbf{Cycle condition}\\
      \hline
      Simple Oscillator (SO) & $\dot{r} = r(a-r^2) \qquad\dot{\theta} = \omega$ & $x_1,x_2\in[-1,1]$ & $a \in [-0.5,0.5]$, $\omega\in [-1,1]$ & $a>0$\\
      \hline
      Li\'enard Polynomial & $\dot{x}_1 = x_2 \qquad\dot{x}_2 = -(a x_1 + x_1^3) -(c + x_1^2) x_2$ & $x_1,x_2\in[-4.2,4.2]$ & $ a\in [0,1]$, $c\in[-1,1]$ & $c<0$\\
      \hline
      Li\'enard Sigmoid & $\dot{x}_1 = x_2 \qquad\dot{x}_2 = -(1 / (1 + e^{-ax_1}) - 0.5) -(b + x_1^2) x_2$ & $x_1,x_2\in[-1.5,1.5]$ & $ a\in [0,1]$, $b\in [-1,1]$ & $b<0$ \\
      \hline
      Van der Pol & $\dot{x}_1 = x_2\qquad\dot{x}_2 = \mu x_2 - x_1 - x_1^2x_2$ & $x_1,x_2\in[-3,3]$ & $\mu \in [-1,1]$ & $\mu >0$\\
      \hline
      BZ Reaction & $\dot{x}_1 = a - x_1 - \frac{4x_1x_2}{1 + x_1^2}\qquad\dot{x}_2 = b x_1 \left(1 - \frac{x_2}{1 + {x_1}^2}\right)$ & $x_1,x_2\in[0,10]$ & $a\in [2,19]$, $b\in [2, 6]$ & $b<\frac{3a}{5}-\frac{25}{a}$\\
      \hline
      Sel\'kov & $\dot{x}_1 = x_1 + ax_2 + x_1^2 x_2\quad\dot{x}_2 = b - a x_2 - x_1^2 x_2$ & $x_1,x_2\in[0,3]$ & $ a\in [0.01,0.11]$, $b\in [0.02,1.2]$ & $b^{-2}>2-4a\pm\sqrt{1-8a}$\\
      \hline
    \end{tabular}
  }
\label{tab:gov-eqs}
\end{table}
We additionally study the \emph{Augmented Simple Oscillator (Augmented SO)} system for the comparison in \Cref{fig:grid-results}, which is an instance of a simple oscillator (SO) whose 2D phase space was transformed by a random initialization of a neural spline flow \cite{durkan2019neural}.
All of these dynamical systems (\Cref{tab:gov-eqs}) undergo a Hopf bifurcation and change their dynamics from exhibiting node attractors to limit cycles, at known parametric settings, making them particularly relevant for our application. 


\subsection{Simulating sparse data}\label{app:sparse-sim}
\begin{figure}
\begin{center}
    \includegraphics[width=\linewidth]{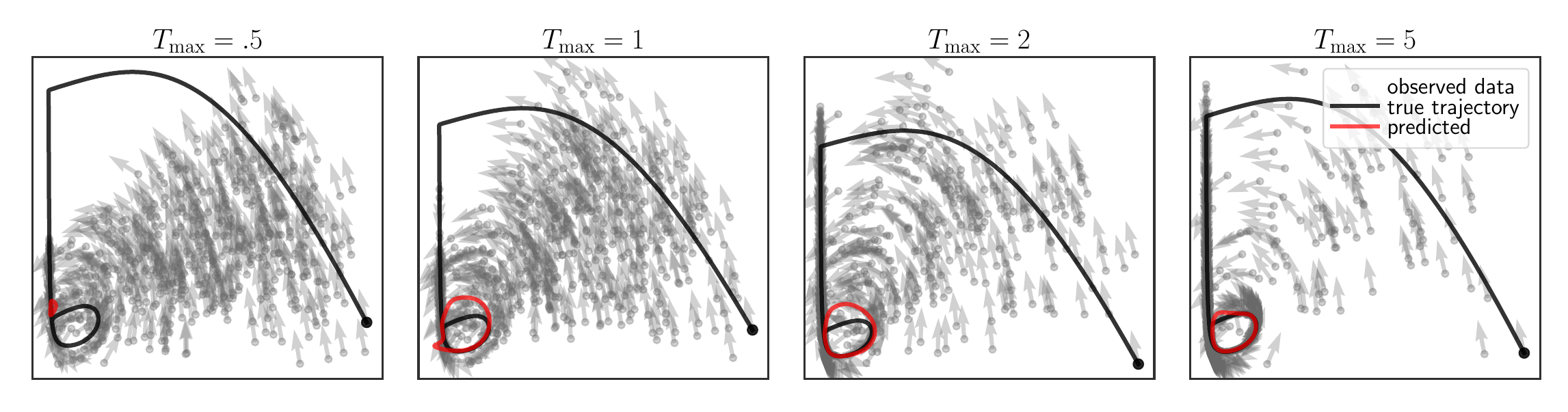}
\end{center}
  \caption{\textbf{The effects of changing the maximum simulation sampling time $T_\text{max}$.} When $T_\text{max}$ is small (e.g. $T_\text{max}=0.5$; left), most of the observations are transients. On the other hand, when $T_\text{max}$ is large (e.g. $T_\text{max}=5$; right), most of the observations are on the invariant set.}
  \label{fig:time-effect}
\end{figure}

A straightforward method to draw sparse data from each dynamical system is to draw observed points uniformly at random from the simulated phase space. However, this process is generally inconsistent with sparse measurements collected in experiments, and for high-dimensional systems this runs the risk of under-sampling the invariant set region, which is a prerequisite for meaningful predictions. 
Instead, 
we resort to the following data sampling process, where each observed pair $(x_i,\dot{x}_i)$ is generated by simulating a trajectory starting from a pre-specified range of the phase-space for a randomly drawn time:
\begin{enumerate}
    \item For each $i\in\{1,\cdots,N\}$, draw the initial conditions $x_i^{(0)}$ uniformly from a pre-specified portion of phase space. 
    \item Sample $t_i$ uniformly in the range $[T_{\min},T_{\max}]$.
    \item Simulate a trajectory beginning at $x_i^{(0)}$ at time $t = 0$ up to time $t_i$, resulting in an observed position and velocity: $x_i^{(t_i)}$ (denoted as $x_i$ for short), $\dot{x}_i=f(x_i)$.
    \item Optionally, Gaussian noise with standard deviation $\sigma$ is then added to $\dot{x}_i$ 
\end{enumerate}

We set the parameter $T_{\min}=0$ for all simulated sparse data in this study (\Cref{fig:invariants,fig:invariant-set-variance,fig:sparse-results}), except for the repressilator, where it is set to $T_{\min}=3$ to ensure that when the samples are projected to lower than six dimensions they still define a vector field with non-intersecting trajectories. 
$T_\text{max}$ is a hyperparameter that effectively controls how much samples tend to concentrate around the invariant set.  
An example of the effect of different values for $T_\text{max}$ can be seen in \Cref{fig:time-effect}. When $T_\text{max}$ is close to 0, the pairs $(x_i,\dot{x}_i)$ revert back to being uniformly sampled from phase-space. When $T_\text{max}$ is small, the observations mostly include transients of the dynamics. As $T_\text{max}$ increases, more points aggregate near the invariant set, which is in turn easier to estimate. For our 2D simulations we use $T_\text{max}=3$, whereas for the repressilator system we used $T_\text{max}=10$. A higher $T_\text{max}$ value was used for the repressilator as trajectories in this higher-dimensional system take longer to converge to the invariant set. 

All trajectories are simulated with the Runka-Kutta method (RK4) and a step size of $\Delta t=0.01$.

\subsection{Repressilator model }\label{app:repressilator-sim}
The repressilator system is governed by six coupled differential equations, structured as three instances of the following gene-regulation model \citep{elowitz2000repressilator}:  
\begin{align*}
    \dot{m_i} &= -m_i + \frac{\alpha}{1 + p_j^n} + \alpha_0,  \\
    \dot{p_i} &= -\beta(p_i - m_i),
\end{align*}
where \( m_i \) and \( p_i \) denote the mRNA and protein concentrations for gene \( i \), corresponding to LacI, TetR, and \(\lambda\) cI, with cyclic inhibition: \( i = \) LacI, TetR, cI and \( j = \) cI, LacI, TetR. The parameters \( \alpha_0 \) (basal transcription), \( \alpha \) (maximal transcription), \( \beta \) (protein/mRNA degradation ratio), and \( n \) (Hill coefficient) define the system's regulatory dynamics.  

We analyze trajectories across \( \alpha \in (0,30) \), \( \beta \in (0,10) \), with \( \alpha_0 = 0.2, n = 2 \). A supercritical Hopf bifurcation emerges at critical values of \( \beta \) \citep{verdugo2018hopf}:  
\begin{align*}
\beta_1 &= \frac{3A^2 - 4A - 8}{4A+8} + \frac{A\sqrt{9A^2 - 24A-48}}{4A+8},\\
\beta_2 &= \frac{3A^2 - 4A - 8}{4A+8} - \frac{A\sqrt{9A^2 - 24A-48}}{4A+8},
\end{align*}
where:
\begin{equation}
    A = \frac{-\alpha n \hat{p}^{(n-1)}}{(1 + \hat{p}^n)^2}, \quad \hat{p} = \frac{\alpha}{1 + \hat{p}^n} + \alpha_0
\end{equation}
These are the boundaries plotted in \Cref{fig:repressilator} (top).


\section{Baseline details}\label{app:baselines}

\subsection{Baselines in invariant set localization}\label{app:inv-baselines}

In \Cref{sec:recon} we showcase the ability of SPE to estimate the position of limit cycles from observations. The few existing approaches that can do so rely on data which is dense on a regular grid \citep{helman1989criticalpts,chen2007morsedesign}. 
Instead, we compare the quality of the fit of SPE to other methods which take as input sparse data and estimate an ODE, and in particular how close limiting structures of the ODEs determined by these different methods approximate the true underlying invariant set.
Accordingly, we compared SPE to a number of baselines: 
\begin{itemize}
    \item \textbf{kNN}, a simple interpolator which uses the $k$-nearest observed positions in order to estimate the velocity. In our experiments, we found that $k=5$ neighbors approximates clean systems fairly well, and used that setting for all of our comparisons.
    \item \textbf{SINDy \citep{brunton2016sindy,deSilva2020pysindy}} is an established method for estimating vector fields from empirical observations. SINDy uses sparse linear regression with a fixed library of functions to estimate the velocities of the system. For our evaluations we use the \texttt{pysindy} package \citep{deSilva2020pysindy} with a function library composed of either 3rd order polynomials (SINDy-poly) or 10 Fourier features (SINDy-fourier). All other hyperparameters were set according to the default values in the \texttt{pysindy} package.
    \item \textbf{MLP} is a simple multi-layer perceptron with 2 hidden layers, with a width of 128 and SiLU activations \citep{hendrycks2016gaussian}. These MLPs were trained to estimate the velocities from the positions using an MSE loss over the predicted velocities, with the Adam optimizer \citep{kingma2014adam}, for 5000 (full-batch) iterations. We used a learning rate and weight decay of $10^{-3}$. Empirically we observed that these hyperparameters result in MLPs that accurately approximate systems from $N=1000$ noiseless observations. 
\end{itemize}
All of these methods (including SPE) take as input the sparse $(x,\dot{x})$ data, and can estimate velocities at new positions, $\dot{x}=f_\theta(x)$, effectively defining an ODE. Using these velocity estimates, it is possible to simulate long trajectories which converge at the fitted system's invariant set. 

\subsection{Evaluation of invariant set localization}\label{app:invariant-eval}

To quantitatively evaluate the localization of the invariant set, we simulate trajectories from initial conditions in both the true (hidden) system and in the fitted systems.
These long trajectories give an effective stationary distribution of points along the invariant set of the considered systems. Accordingly, our goodness-of-fit measure is the Wasserstein-2 ($W_2$) distance between effective stationary distributions of the hidden governing equations and those generated by the estimated velocities for each model.

For the evaluations in \Cref{fig:sparse-results} (left), we randomly sampled 1000 systems which contain limit cycle behaviors, uniformly from the set containing SO, Li\'enard Polyonmial, Li\'enard Sigmoid, Van der Pol, BZ Reaction and Sel\`kov families of systems (\Cref{app:2D-systems}), with order parameters uniformly sampled from the ranges corresponding to limit cycle behaviors. For each system, different numbers $N$ of positions $x_i$ and velocities $\dot{x}_i$ were sampled according to the procedure described in \Cref{app:sparse-sim}, with $T_\text{max}=3$. Additive Gaussian noise was then added to these velocities with a standard deviation of $\sigma = \mu/\text{SNR}$ where $\mu=\frac{1}{N}\sum_i \|\dot{x}_i\|$ and a prespecified signal-to-noise ratio (SNR). Adding noise in this manner allows for direct control of the SNR of the observed velocities between different families of systems, whose domains are defined at different scales. Each of the methods, SPE and the baselines described in \Cref{app:inv-baselines}, were then fitted to all $N$ simulated positions $x_i$ and noisy velocities $\dot{x}_i$.

Each method was then evaluated by comparing their effective stationary distribution to that of the ground-truth system. To do so, $M=1000$ initial conditions were sampled from the range of initial conditions of the corresponding family of systems (see \Cref{app:2D-systems}). The same initial conditions were used to generate long trajectories under the true system and the one determined by the fitted velocities, and the $W_2$ distance was then calculated between the last time-points for the true system and the fitted ODE. The trajectories were integrated using the Runka-Kutta method, with a step size of $\Delta t=0.01$, for $T_\text{eval}=100$ which we found to be long enough for the true system to be close to its invariant set.
To calculate the $W_2$ distance we use the Python Optimal Package from \citet{flamary2021pot}. 

\subsection{Evaluation given prototype mismatch}\label{app:mismatch-details}

To evaluate the effects of prototype mismatch, when the input prototype is inconsistent with the dynamical systems underlying the observed data, we considered a scenario where both prototypes and data exhibit limit cycles with either clockwise (CW) or counter-clockwise (CCW) angular motion. Because our parametric diffeomorphisms (\Cref{sec:nf-diffeos}) cannot change the orientation of the motion, this results in a potential mismatch: if the data contains CW motion but the prototype has CCW motion (or vice-versa), then the two are mismatched. Otherwise, the data and prototype are considered equivalent. For CW based systems, we used the simple oscillator (SO) with $\omega <0$, Sel\'kov, Li\'enard sigmoid, Li\'enard polynomial and Van der Pol family of systems from \Cref{app:2D-systems}. For systems with CCW motion, we used SO with $\omega>0$ and the BZ reaction family of systems. The prototypes in these experiments have the same settings as in \Cref{sec:recon} - an SO system with $a=0.25$ and $\omega=0.5$ for CCW motion and $\omega=-0.5$ for CW angular velocity. All trained networks used 4 blocks with a width of 5, were trained for 500 iterations with a learning rate and weight decay of $10^{-3}$. 

We observed that when there is a mismatch between data and prototype, the diffeomorphism tends to overfit and can change drastically under small perturbations of the data.
To quantify these effects, we fit 5 different diffeomorphisms to 5 random splits of the data, where 50\% of the data points are randomly omitted. We then calculated the variance in the position of the inferred localized invariant set under the different splits. 

\subsection{Benchmarking classification results}\label{app:classif-baselines}

To benchmark SPE in the classification of dynamical systems, we turn to datasets and existing baselines previously considered by \citet{moriel2023let}. 
Below we provide a brief summary of each of these methods: 
\begin{itemize}
    \item \textbf{TWA} \citep{moriel2023let}, a convolutional NN trained to classify between point and periodic dynamics on a dense grid. Instead of training on the input vectors directly, TWA predicts the class of the system using an angular representation of each vector in each position. TWA is trained on simulated data, which is generated by sampling the order parameters from the SO family of systems. These are then augmented using neural spline flows \citep{durkan2019splineflow}. Augmenting in this way ensures that the same invariant structure is kept between the originally sampled SO system and the augmented version. TWA was then trained on 10,000 of these randomly augmented systems.
    \item \textbf{Critical Points}, is based on an algorithm devised by \citet{helman1989criticalpts} which detects and characterizes critical points from a representation of an autoencoder. A system is considered periodic under CriticalPoints if at least one detected point is a repeler, and is otherwise classified as an attractor. 
    \item \textbf{Phase2vec} \cite{ricci2022phase2vec} is a convolutional autoencoder trained to reconstruct 10,000 training systems defined by sparse polynomial equations. The encoder maps to a 100-dimensional latent space using a single layer. The outputs of the decoder are the predicted coefficients for the sparse polynomial equations. These coefficients are then multiplied by the polynomial function library to reconstruct the original vector field. The latent representations of the autoencoder then form features for an additional linear classifier, which was trained for classification using the same dataset as TWA.
    \item \textbf{Autoencoder} is a convolutional autoencoder with a bottleneck of 10 latent dimensions, trained to directly reconstruct the vectors generated as in Phase2vec. Both encoder and decoder have three convolutional blocks with a stride of $2\times 2$ to reduce the resolution of the input during training. After training, similarly to Phase2vec, the latent representations are turned into features for a linear classifier fitted to the same data as TWA.
\end{itemize}

To benchmark SPE against existing methods for classification between node or cyclic behaviors, we use the dataset introduced by \citet{moriel2023let}. This dataset is based on the simulated families of systems with order parameters corresponding to either node or periodic attractor, described in \Cref{app:2D-systems}. Accordingly, the task is to classify each system in the dataset as either a point or periodic attractor. Systems in the dataset are comprised of a set of positions $x_i\in\mathbb{R}^2$ and the velocities corresponding to those positions, $\dot{x}_i\in\mathbb{R}^2$, organized on a dense, regular $64\times64$ grid for a total of $N=4096$ observations. 
\Cref{fig:grid-results} shows the average accuracy of SPE for each family of systems, compared to existing baselines.
SPE outperforms existing baselines across all families of systems except for the family of simple oscillators (SO) and Augmented SO, where TWA, which was trained on thousands of these systems using a dense grid of points in phase space, exhibits higher accuracy (\Cref{fig:grid-results}). 

Note that, unlike most of the available baselines, SPE is not restricted to systems defined on a regular grid and can directly localize the invariant set (as shown in \Cref{sec:recon}).

\begin{table*}
\centering
\caption{\textbf{Comparison of the classification accuracy of SPE to different models across various families of dynamical systems, when classifying between periodic and node dynamics.} The classification accuracy is averaged over 100 instanced for each family of system (see \Cref{app:2D-systems} for details). The observed vectors were positioned on a dense $64\times64$ grid, with the addition of Gaussian noise with a standard deviation of $\sigma=0.1$ on the velocities. Where relevant, the errors in accuracy are one standard deviation, calculated over different initializations.
The highest scores for each column are shown in bold, with the second highest underlined.}
\resizebox{\textwidth}{!}{
\begin{tabularx}{\linewidth}{@{}lXXXXXXXX@{}}
\toprule
           & SO            & Aug. SO       & Li\'enard Poly  & Li\'enard Sigmoid & Van der Pol   & BZ Reaction   & Sel\'kov  & Attractor estimation\\ \midrule
SPE  &      91±1            & 81±1    & \textbf{96±1}  & \textbf{92±1}  & \textbf{97±3}  & \textbf{88±2}  & \textbf{68±1}   & \checkmark           \\
TWA  & \textbf{98±1} & \textbf{93±1} & \underline{86±13} & \textbf{92±7}    & 83±15 & 82±11 & \underline{65±3} & \text{\sffamily x}\\
Critical Points & 54          & 56          & 70          & 49              & 56          & \underline{84}          & 51        & \text{\sffamily x} \\
Phase2vec   & 73±8     & 71±4     & 49±6     & 48±0.1          & 49±4     & 50±0.1      & 49±0.1      & \text{\sffamily x}\\
Autoencoder & \underline{95±3}     & \underline{87±1}     & 63±16     & 88±9         & 81±16     & 82±13     & 49±2    & \text{\sffamily x} \\ \bottomrule
\end{tabularx}
}
\label{fig:grid-results}
\end{table*}

\subsection{Evaluation in sparse classification}\label{app:sparse-classif-eval}

To evaluate the classification performance of SPE on sparse and noisy data, we simulated systems in the same manner as in \Cref{app:invariant-eval}. We randomly sampled 1000 systems from the SO, Li\'enard Polyonmial, Li\'enard Sigmoid, Van der Pol, BZ Reaction and Sel\`kov families of systems. For each system, the values of the order parameters were uniformly sampled from their corresponding ranges (see \Cref{app:2D-systems}), which correspond to either node attractors or cyclic attractors. For each system, $N$ positions $x_i$ and velocities $\dot{x}_i$ were sampled according to the procedure described in \Cref{app:sparse-sim}, with $T_\text{max}=3$. Additive Gaussian noise was then added to these velocities with a standard deviation of $\sigma = \mu/\text{SNR}$ where $\mu=\frac{1}{N}\sum_i \|\dot{x}_i\|$ and a prespecified signal-to-noise ration (SNR). The task was then to classify the set of pairs $\left\{(x_i,\dot{x}_i)\right\}_{i=1}^N$ either as a node attractor or a limit cycle.

\section{Identification of the cell cycle trajectory from single-cell RNA-sequencing and RNA velocity data of proliferating cells}\label{app:fucci}

For the scRNA-seq data (\Cref{sec:fucci}), we use two datasets of proliferating cell lines: U2OS cell line data collected in \citet{mahdessian2021fuccicycle} and human fibroblast cells from \citet{riba2022cycle}. In both cases, the data is predominantly comprised of cells that are proliferating. The U2OS cell line data includes 1,152 cells characterized by the expression of 58,884 genes, whereas the human fibroblast data contains 3,086 cells and 10,789 genes. For preprocessing, we followed the protocol in \citep{zheng2023pumping}, which involved gene and cell filtering, normalization, log transformation, and the selection of highly variable genes. We also use labels provided in both studies to filter out cells which are not proliferating. 

To infer local measures of cellular dynamics in the high-dimensional gene expression space, we used scVelo \citep{bergen2020scvelo} which computes RNA velocity vectors \cite{la2018rna} from spliced and unspliced gene expression counts of a cell population. scVelo models the transcriptional dynamics of each gene using a dynamical model of transcription, splicing, and degradation. It applies an Expectation-Maximization algorithm to estimate gene-specific kinetic parameters for each cell on smoothed versions of the spliced and unspliced counts, which are estimated from a $k$-nearest neighbor graph over gene expression \citep{bergen2020scvelo}. 

The gene expression data and the inferred velocities were projected into a lower dimensional space using principal component analysis (PCA), of dimensionality 100 for the U2OS cell line and 10 for the human fibroblasts. 
In both cases, using PCA greatly reduces the computational load of our method, while retaining almost all variability in the data, and allowing us to project back into gene expression space after fitting.
The option to project back into gene expression space enables the projection of invariant sets found using SPE into the full gene expression space and to probe the specific role of each gene in trajectories on the invariant set.
Streamlines of the resulting velocity field are visualized in \Cref{fig:fucci}, left. Independently of the RNA velocities, we computed S-phase and G2/M-phase scores per cell by aggregating the expression of marker genes for these phases, as defined by \citet{tirosh2016dissecting}, using the function ``score\_genes\_cell\_cycle'' from scVelo package \citep{bergen2020scvelo}. The color in \Cref{fig:fucci}, left and middle, represents the ratio of the S-phase score (red) and the G2/M-phase score (blue).

We then applied our method to fit the RNA velocities, embedded in the PCA space, to a high dimensional cyclic oscillator prototype, as defined in \Cref{eq:multidim} of \Cref{sec:highdim-prototypes}. This prototype oscillates in the first two dimensions and decays exponentially in the others. The fitting parameters were as follows: angular speed $\omega = 5$, radius $a = 1$, three blocks as defined in \Cref{app:diff-details}, $K = 3$ frequencies per layer, weight decay of $\texttt{wd}=10^{-3}$, projection regularization set to $\texttt{proj\_reg}=e^{-2}$, determinant regularization of 0, a learning rate of $\texttt{lr}=5\cdot 10^{-4}$, and 1,000 iterations.

Given the fitted coordinates of the cycle attractor in PCA space, we inverted the PCA transformation to find the expression of the different genes along the fitted cycle. 
These predicted gene expressions were normalized to the range [0,1], by subtracting the minimal value and then dividing by the maximal value, for visual clarity.
In \Cref{fig:fucci} (right), we plot the normalized gene expression of representative marker genes for both S-phase 
and G2/M-phase 
(\citet{tirosh2016dissecting}), showing distinct oscillatory patterns consistent with cell cycle progression.

\section{Further experiments}

\subsection{Multiple basins of attraction}\label{app:multistable}

\begin{figure*}[!tbh]
\begin{center}
    \includegraphics[width=\linewidth]{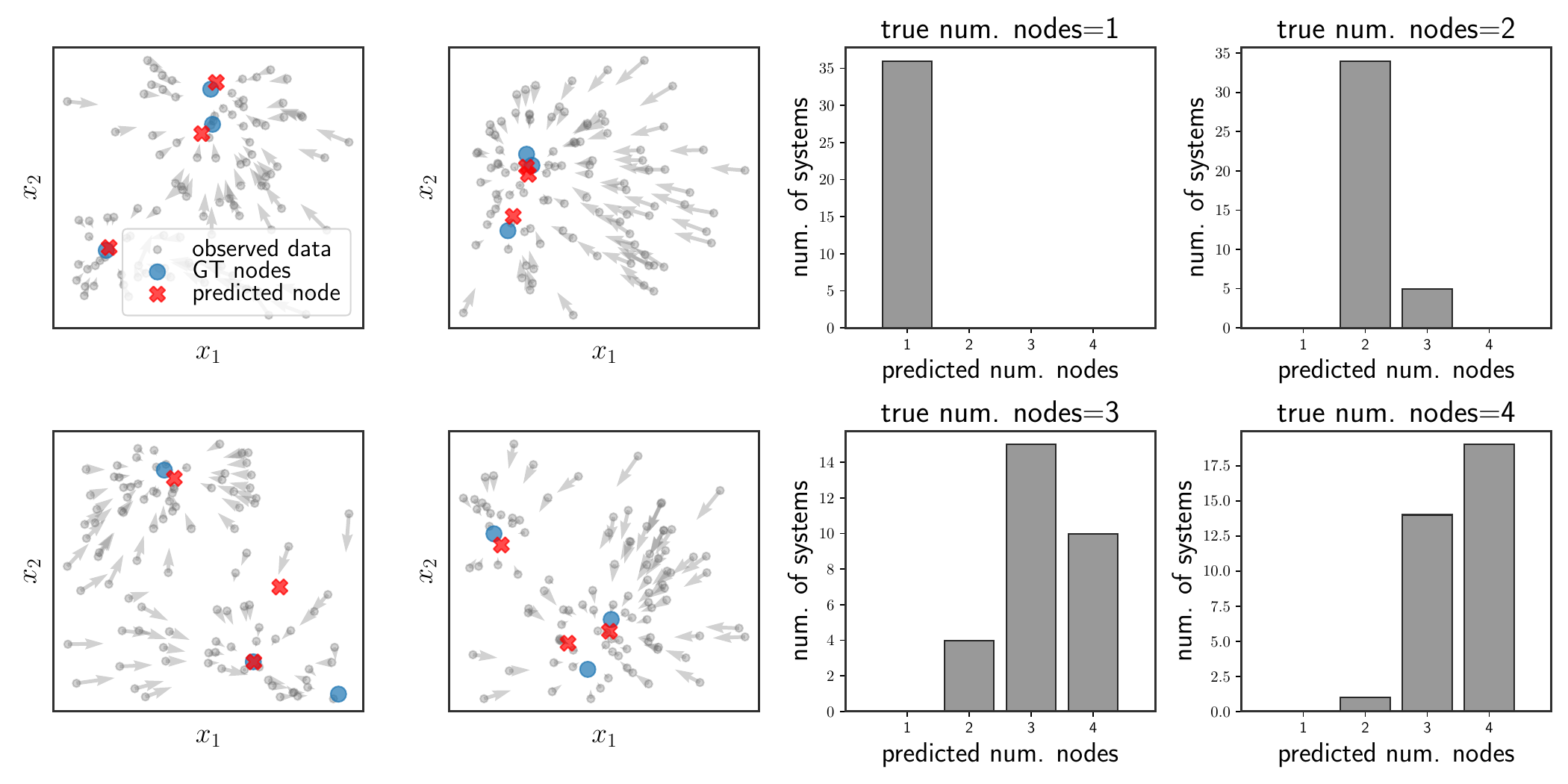}
\end{center}
  \caption{\textbf{Classifying and localizing invariant sets in systems with multiple basins of attraction.}
  \textbf{Left:} 4 examples of fitting SPE to a system with multiple basins of attraction in 3 dimensions. In each example, the positions of the node attractors was randomly sampled (these positions are depicted by the blue circles in the plot), and 100 position-velocity pairs were then sampled using the scheme described in \Cref{app:sparse-sim}. SPE was then used to match between this data and a fixed prototype with 3 basins of attraction. The estimated positions of the attractors according to SPE is depicted by the red crosses. We find that SPE frequently matches the positions of the nodes.
  \textbf{Right:} results in the classification of the number of nodes. 
  Each bar plot shows how SPE categorizes systems with a set number of nodes, for instance the top left plot shows the performance only on systems that had one node attractor. In each of these cases, the main mode predicted by SPE aligns with the true number of nodes in the system.
  }
  \label{fig:multistable}
\end{figure*}

While this work mostly focused on prototypes with a single basin of attraction, such as a node attractor or a limit cycle, SPE can also be used given multiple basins of attraction. 

As a simple example, we studied synthetic dynamics with multiple fixed points in a high-dimensional space. The governing equations for these dynamics were based on a mixture of experts decomposition of space, where each expert represents a single basin of attraction. Specifically, the governing equations of the dynamics were defined as:
\begin{align}
    f\left(x\right)&=\sum_{k=1}^K r_k\left(x\right) \varphi_k\left(x-c_k\right)\\
    r_k\left(x\right)&=\frac{\exp\lrbra{-\frac{1}{2 s^2}\|x-c_k\|^2}}{\sum_{k'}\exp\lrbra{-\frac{1}{2 s^2}\|x-c_{k'}\|^2}}
\end{align}
where $\varphi_k(\cdot)$ are the dynamics of expert $k$, whose coordinate system is centered around $c_k\in\mathbb{R}^d$, $s$ determines the size of the region that the $k$-th expert models,
and $K$ is the number of basins of attraction. Through these, $f(x)$ models dynamics across the whole space, where for each expert $0< r_k(x)<1$ determines how strongly the dynamics at position $x$ are controlled by expert $k$.

Synthetic data was generated by sampling systems with centers sampled uniformly from the range $\lrbra{-0.9,0.9}^d$, all of which were node attractors:
\begin{equation}
    \varphi_k(y)=-y
\end{equation}
For each generated system, the number of attractors $K$ was chosen between 1 and 4. From these generated systems, we followed the sampling procedure from \Cref{app:sparse-sim} to create 100 pairs of positions and velocities. In our experiments, we considered 5-dimensional data, meaning that each node $c_k$ is determined by a 5-dimensional vector.

We then fit SPE to each system generated this way, with prototypes of the same form
- 4 prototypes, one for each possible number of centers. This allowed us, in turn, to classify the number of basins of attraction in the observed data.
The positions of the nodes $c_k$ in the prototypes were fixed and distinct from the positions of the nodes in the true underlying system, so that the fitting process decomposes into the prediction of both the number of nodes as well as their location. Results can be seen in \Cref{fig:multistable}. 


For the experiments shown in \Cref{fig:multistable}, we used INNs with 2 blocks of width 3, and 5 Householder transformations. The networks were trained for 1500 iterations, with a learning rate of $1e^{-3}$, weight decay of 0.1, with all other regularization coefficients equal to 0. 

\subsection{Mismatch between prototype and underlying system}\label{app:mismatch}
\begin{figure*}[!tbh]
\begin{center}
    \includegraphics[width=\linewidth]{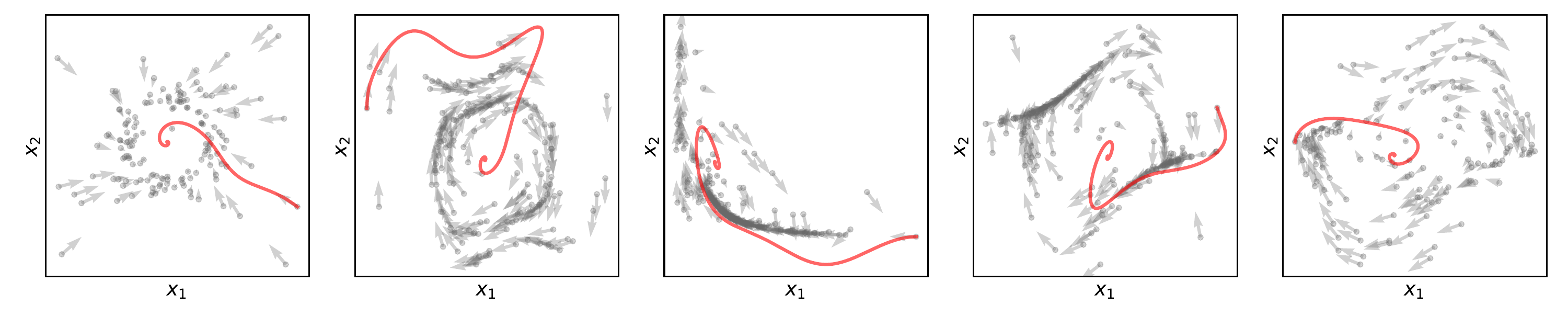}
\end{center}
  \caption{\textbf{Direction of angular velocity is captured, even when there is a mismatch between the prototype and the true underlying system.} Randomly sampled systems and examples of fitting SPE to an inherently oscillatory process using a node attractor prototype. Despite this mismatch between underlying behavior and prototype, SPE is still able to characterize the dominant behavior of the system, mainly the direction of the angular velocity and the focal point of the limit cycle.
    }
  \label{fig:mismatch}
\end{figure*}

SPE uses a soft notion of equivalence to match between the prototypes and the data. Thus, even when there is a mismatch between the prototype and the underlying system, SPE converges to a solution that transforms the dynamics of the chosen prototype to resemble the observed dynamics. 
Because of this, we expect SPE to find a mapping that retains certain characteristics of the dynamics that are shared between the data and the prototype, even in cases where the two systems are not fully smoothly equivalent. 
We tested this through the analysis of simulated oscillatory systems from \Cref{app:2D-systems}. 
Systems representing node attractors and limit cycles, while not equivalent, can still have shared attributes. One example is the orientation of the flow - either clockwise or counter-clockwise - around the basin of attraction. 
To emulate a mismatch between the input prototype and the underlying dynamics, we fit SPE using only node attractor prototypes, as opposed to the true underlying oscillatory behavior, using the same prototypes from \Cref{eq:prot}, with $a=-0.25$ and $\omega=\pm0.5$. 
\Cref{fig:mismatch} demonstrates that even when there is a mismatch between the underlying dynamics and the prototype, SPE is able to match the orientation of the flow. Specifically, if the system exhibits a clockwise limit cycle, SPE fits a node attractor spiraling clockwise, corresponding to a negative value for the angular velocity parameter, $\omega$, of the prototype, and vice versa. 
Quantitatively, SPE achieves 98\% accuracy in predicting the flow orientation, $\omega$, of the SO family of systems, averaged over 100 trials using the same hyperparameters described in \Cref{app:hyperparams}. 


\subsection{Alternative limit cycle prototypes}

\begin{figure*}[!tbh]
\begin{center}
    \includegraphics[width=.8\linewidth]{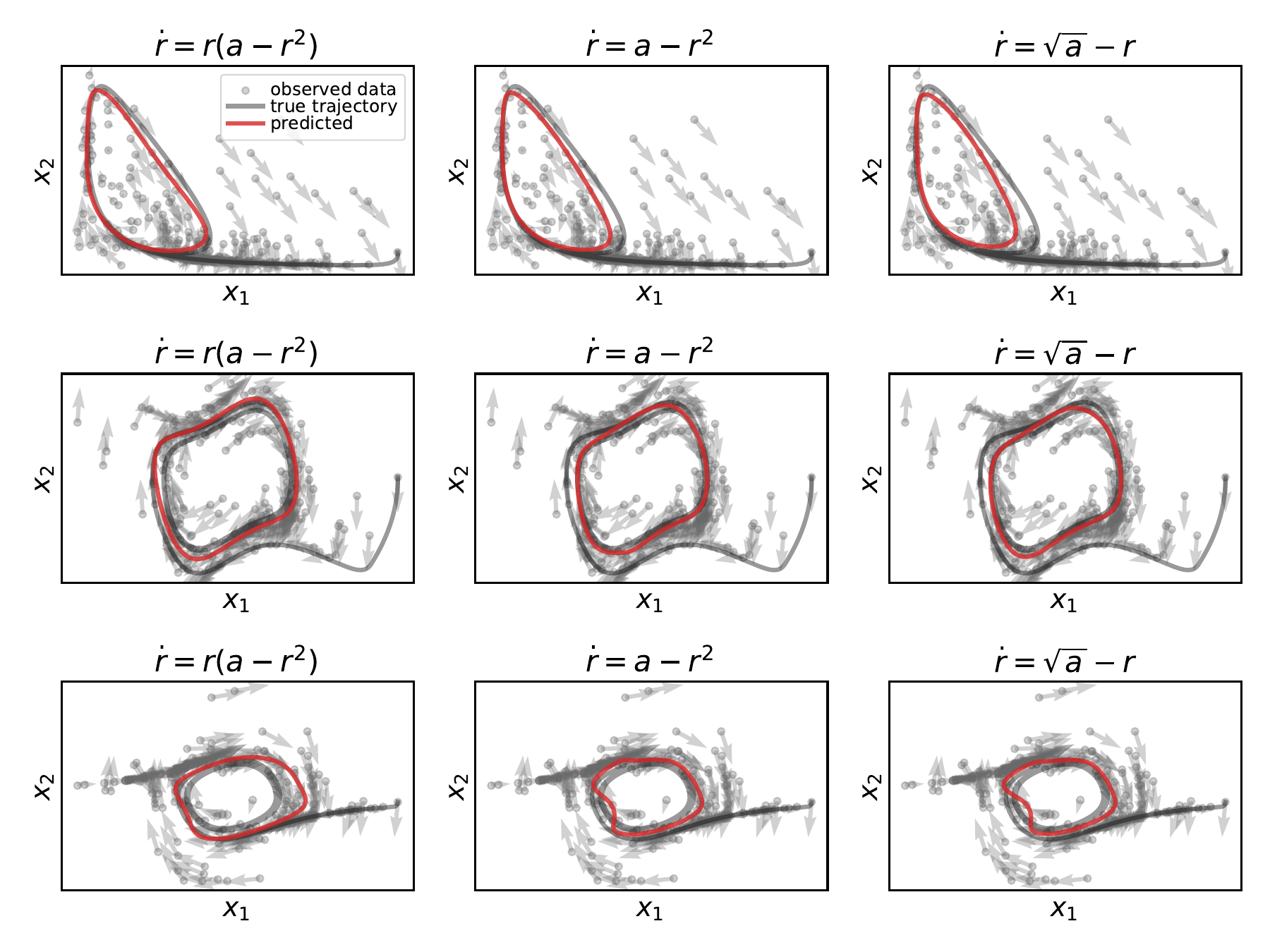}
\end{center}
  \caption{\textbf{Estimation of limit cycles using different oscillator prototypes.}
  The three rows correspond to different possible cycle prototypes. The left-most column corresponds with the prototypes used in the rest of this work, and the middle and right-most columns have decreasing speeds towards the limit cycle in the radial direction. The three prototypes result in similar estimated invariant sets.}
  \label{fig:cycle-protots}
\end{figure*}

Throughout the main text, the form of the prototypes we used were those of the simple oscillator (SO):
\begin{equation}\label{eq:app-hopf}
    \dot{r} = r(a - r^2),\qquad
    \dot{\theta} = \omega
\end{equation} 
where $r$ and $\theta$ are the polar coordinates. We focused on the SO as it can be fully characterized analytically and exhibits a well-studied form of a supercritical Hopf bifurcation, transitioning between a fixed point to a limit cycle attractor. However, SPE shows robust performance of invariant set localization over observations from other forms of dynamical systems exhibiting limit cycle attractors, as these systems are smoothly equivalent to each other. 

This is demonstrated over two alternative prototypes, representing dynamical systems of other oscillator forms: 
\begin{align}
    \dot{r} = a - r^2,&\qquad
    \dot{\theta} = \omega\\
    \dot{r} = \sqrt{a} - r,&\qquad
    \dot{\theta} = \omega
\end{align}\label{eq:alt-cycles}
where $r$ and $\theta$ are the polar coordinates. Here, as in the SO prototype, the parameter $a$ represents the radius of the limit cycle and $\omega$ represents the angular velocity. These systems exhibit different speeds in the radial direction towards the limit cycle, but otherwise remain qualitatively similar. Using SPE for invariant set localization with these alternative prototypes \Cref{eq:alt-cycles}, instead of the SO prototype \Cref{eq:app-hopf}, for the localization of limit cycles in simulated 2D data (systems from \Cref{tab:gov-eqs}) results in similar estimated invariant sets (\Cref{fig:cycle-protots}).




\bibliographystyle{unsrtnat}
\bibliography{biblio}

\end{document}